%% file: main.tex
\documentclass[runningheads]{llncs}

\usepackage{eccv}

\usepackage{eccvabbrv}

\usepackage{graphicx}
\usepackage{booktabs}
\usepackage{multirow}
\usepackage{multicol}
\usepackage{dsfont}

\usepackage[accsupp]{axessibility}  % Improves PDF readability for those with disabilities.

\usepackage{hyperref}
\usepackage{orcidlink}

\usepackage[most]{tcolorbox}

\begin{document}
\newtcolorbox{promptbox}[1][]{
  colback=gray!5!white,      % Background color
  colframe=gray!75!black,    % Border color
  fonttitle=\bfseries,       % Bold title
  title={System Prompt},     % Default Title
  #1                         % Allow optional arguments to override defaults
}
% ---------------------------------------------------------------

\title{Towards Robustness against Typographic Attack with Training-free Concept Localization\texorpdfstring{\thanks{Accepted to the European Conference on Computer Vision (ECCV) 2026.}}{}}

\titlerunning{Training-free Concept Localization against Typographic Attack}

\author{Bohan Liu\inst{1}\orcidlink{0009-0008-6084-9133} \and
Wenqian Ye\inst{1}\orcidlink{0000-0002-6069-5153} \and
Guangzhi Xiong\inst{1}\orcidlink{0000-0002-8049-5298}
\and
Zhenghao He\inst{1}\orcidlink{0009-0004-1518-9747}
\and
Sanchit Sinha\inst{1}\orcidlink{0000-0003-2650-0612}
\and
Aidong Zhang\inst{1}\orcidlink{0000-0001-9723-3246}}

\authorrunning{B.~Liu et al.}

\institute{$^{1}$University of Virginia, Charlottesville VA 22903, USA\\
\email{\{bohan, aidong\}@virginia.edu}}

\maketitle
\input{sec/0_abs}
\input{sec/1_intro}
\input{sec/2_related}
\input{sec/3_method}
\input{sec/4_experiments}
\input{sec/5_discussion}
\clearpage

% \section*{Acknowledgements}
% Please insert your acknowledgments here.


\bibliographystyle{splncs04}
\bibliography{main}

\clearpage
\input{sec/a_appendix}
\end{document}

%% file: sec/0_abs.tex
\begin{abstract}
  Models trained via Contrastive Language-Image Pretraining (CLIP) serve as the foundational vision encoders for many modern Large Vision Language Models (LVLMs). Despite their widespread adoption, CLIP models exhibit a critical yet underexplored failure mode: irrelevant text appearing within images confounds visual representations, biasing them toward lexical meaning rather than true visual semantics. This robustness issue, commonly described as a Typographic Attack (TA), exposes a vulnerability that poses a significant risk to safety-critical applications such as autonomous driving. To achieve interpretable and effective robustness against TA, we propose a novel, training-free mechanistic interpretability method.  Our method provides sampling-based interpretations of hidden state representations and quantitatively attributes semantic versus lexical focus to individual attention heads. Through probabilistic analysis and circuit mining, we isolate specific Vision Transformer (ViT) components that disproportionately encode lexical information, thereby identifying the mechanistic source of TA. We further show that simple interventions applied directly to the identified circuits, without any additional training, can substantially improve robustness against Typographic Attacks in object classification. These interventions, such as selective adjustment of attention weights, outperform both supervised and training-free defense methods. Our experiments also demonstrate that applying the proposed intervention to the vision encoders of several state-of-the-art LVLMs yields substantial gains in Visual Question Answering accuracy under Typographic Attack interference on RIO-Bench. These results confirm both the efficacy and the generalizability of our mechanistic approach. Code is released at \url{https://github.com/Liu-524/SamplingTAR}.
  \keywords{Typographic Attack \and Vision Language Models \and Mechanistic Interpretability}
\end{abstract}

%% file: sec/1_intro.tex
\section{Introduction}
Contrastive Language-Image Pretraining (CLIP) \cite{radford2021learning} has become the cornerstone of modern visual perception systems. The simple contrastive objective of CLIP enables large-scale parametric learning and the alignment of vision and language. CLIP's strong out-of-distribution generalization enables its use as a zero-shot backbone for diverse visual recognition and perception tasks. Consequently, CLIP vision encoders have powered several state-of-the-art Large Vision Language Models~(LVLMs), including LLaVA~\cite{liu2023visual}, Qwen-VL~\cite{bai2023qwen}, and InternVL~\cite{chen2024intern}.

Despite their critical role in increasingly high-stakes LVLM applications~\cite{wang2025surgicallvlm, yang2023survey}, our understanding of the internal mechanism of CLIP models remains limited. Given that most CLIP variants are fundamentally grounded in the Vision Transformer (ViT) architecture~\cite{dosovitskiy2021an}, investigating the ViT encoder is a central imperative~\cite{tong2024eyes}. Previous interpretability efforts have focused on residual stream decomposition~\cite{gandelsman2024interpreting, gandelsman2024neurons} or hidden space disentanglement with Sparse Dictionary Learning~(SDL)~\cite{huben2024sparse, zaigrajew2025interpreting, dreyer2025attributing, gandelsman2024interpreting}. However, these approaches largely analyze the network at a macro level, treating layers as indivisible blocks. Few studies have directly probed the internal attention mechanisms of ViTs to uncover exactly how individual visual patches interact.

While highly capable of aligning vision and language, CLIP models have been shown to exhibit a distinct vulnerability, namely Typographic Attacks~(TA)~\cite{goh2021multimodal}. In contrast to conventional imperceptible adversarial perturbations, typographic attacks exploit the model's recognition capacity by imposing deceptive text onto an image. For instance, an image of the class "cat" with the injected word "Goose" usually leads to misclassification (see \cref{fig:schematic}a). This vulnerability highlights a critical feature entanglement in the vision encoders, in which the encoders cannot effectively decouple the visual features from the semantics encoded in lexical shapes. Therefore, resolving this issue requires methods beyond standard black-box defenses. This motivates us to propose a mechanistic method to precisely determine where and how typographic features arise and interfere with visual semantic processing during encoding.

A recent study on the Linear Representation Hypothesis~\cite{park2024linear, elhage2022superposition} highlights that latent representations can be interpreted as linear combinations of concept directions. Building on this hypothesis, we introduce the stochastic sampling of pseudo-concept vectors as a training-free approach to interpret transformer modules. Additionally, we propose an attribution-based lexical-focus ranking mechanism to identify TA-vulnerable concept samples and pinpoint their host modules. We perform a systematic analysis of stochastic concept mining and leverage dimensionality reduction in Multi-Head Self-Attention~(MHSA) projections to reduce the search space and mitigate concept entanglement. The resulting attribution-based label-free circuit mining method provides faithful and explainable TA robustness to CLIP ViT and LVLMs through simple and fast test-time circuit intervention. We summarize our contributions as follows:
\begin{itemize}
\item \textbf{Sampling-based Concept Mining.} Building on the Linear Representation Hypothesis~\cite{park2024linear, elhage2022superposition} and Sparse Autoencoders, we provide a training-free stochastic sampling method of concept mining and demonstrate that leveraging the natural decomposition of representation in Multi-Head Self-Attention modules provides a significant increase in the likelihood of a concept hit.
\item \textbf{Model Vulnerability Explanation.} We propose a gradient-based attribution method that jointly considers the attention head "focus" and the concept "direction" to obtain a normalized attribution score that consistently identifies lexical-focusing concepts and attention modules.
\item \textbf{Mechanistic Intervention for Improved Robustness.} Attention reweighting and ablation on harmful circuits significantly improve CLIP's object classification accuracy and LVLM's Visual Question-Answering~(VQA) accuracy on typographic attack datasets. Our method is shown to outperform prior state-of-the-art methods and improve large LVLMs at minimal cost. 
\end{itemize}

%% file: sec/2_related.tex
\section{Related Works}
\label{sec:related}
\subsection{Interpretability on Attention Heads}
Multi-Head Self Attention (MHSA)~\cite{vaswani2017attention} has become a fundamental building block of Transformers. Conceptually, MHSA creates parallel, lower-dimensional views of the same input data, applies separate attention operations to each view, and combines the views across heads. Despite its widespread adoption, the interaction between attention heads and their distinct behaviors remains underexplored. \cite{li-etal-2023-interpreting} applies attention-head pruning to reveal the importance of heads for specific tasks in a multi-task learning setup and exploits head-task affinity to improve multi-task learning. Specifically, in the context of CLIP model interpretability, \cite{gandelsman2024interpreting} studies the decomposition of heads and layers in CLIP-ViT models by doing a greedy search on a set of text descriptions that maximizes the explained variance of each head's latent space. More recently, \cite{su2025concepts} exploits the output and value projections to the residual stream~\cite{elhage2021mathematical} of transformers to directly analyze the attention head's concept vectors in a learned dictionary. \cite{kissane2024interpretingattentionlayeroutputs} extends self-supervised interpretation by directly applying a Sparse Dictionary Learning~(SDL) to the attention layer outputs, with attributes based on the norms of the decoder channels. Based on the SDL intuition and the Linear Representation Hypothesis~\cite{park2024linear, elhage2022superposition}, our work investigates stochastic sampling as a model interpretability tool when combined with mechanistic attribution and reduces the computational and data costs of parametric dictionary learning. Our method tailors the model interpretation process to the ViT architecture by jointly considering attention weighting and concept alignment and provides faithful attribution to TA-vulnerable modules based on the spatial information flow within ViT. 
\subsection{Typographic-Attack Robustness}
Typographic-Attack Robustness has received increasing attention as the development and application of complex LVLMs have advanced, which rely on robust CLIP visual embeddings. Evaluations on TypoD~\cite{unveil} confirm the prevalence of TA vulnerability in CLIP ViT and LVLMs. \cite{wang-etal-2025-typographic} extends Typographic Attack to a multi-image setting.
\cite{chung2024towards} applies typographic attack techniques to self-driving scenarios, and SceneTap~\cite{scenetap} extends the scope of Typographic Attack in self-driving to the occurrence of realistic and coherent text distractors in images. Shown in \cref{tab:vit_comparison}, on the defense side, Defense-Prefix tuning~\cite{denseprefix} applies prompt token learning on annotated data to trigger lexical overlook behavior of CLIP models. Recently, advances in TA robustness~\cite{dyslexify} have shifted toward explainable, training-free interventions for vulnerable modules. Our work proposes optimization-free stochastic sampling as a replacement for expensive parametric dictionary learning and pushes the boundary of training-free intervention for TA robustness.
\begin{table}[tb]
\centering
\caption{Comparison of defense properties within the ViT paradigm. Our method leverages the specific architectural priors of Transformers to achieve robustness without requiring labeled greedy searches.}
\label{tab:vit_comparison}
\resizebox{1\textwidth}{!}{
\setlength{\tabcolsep}{10pt}
\begin{tabular}{@{}lcccc@{}}
\toprule
\textbf{Method} & \multicolumn{1}{|c}{\textbf{Training-Free}} & \multicolumn{1}{|c}{\textbf{Low Data}} & \multicolumn{1}{|c}{\textbf{Interpretable}} & \multicolumn{1}{|c}{\textbf{Intervention Cost}} \\ \midrule
Defense-Prefix~\cite{denseprefix} & $\times$ & $\times$ & $\times$ & High \\
Dyslexify~\cite{dyslexify} & $\checkmark$ & $\times$ & $\checkmark$ & High (Iterative) \\ \midrule
\textbf{Ours} & $\checkmark$ & $\checkmark$ & $\checkmark$ & \textbf{Low (Constant)} \\ \bottomrule
\end{tabular}
}
\end{table}

%% file: sec/3_method.tex
\section{Method}
\begin{figure}[t]
    \centering
    \includegraphics[width=0.95\linewidth]{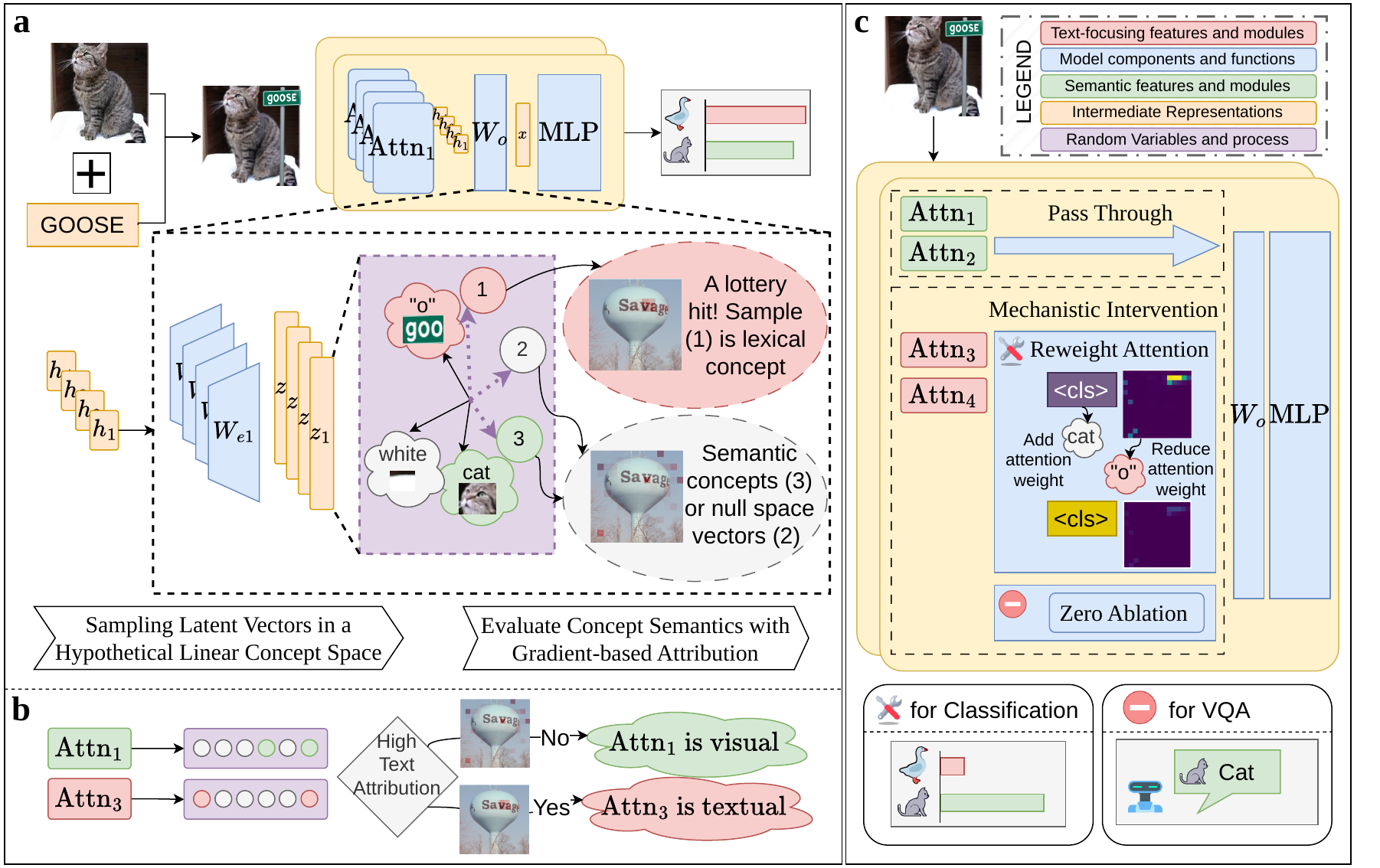}
    \caption{\textbf{An overview of the proposed method.} \textbf{a)} shows the sampling-based mechanistic interpretability (the \textbf{Stochastic Lottery}) for lexical circuit mining. Latent samples (denoted by purple dashed arrows) on the hypothesized concept basis reveal distinct attribution patterns. \textbf{b)} illustrates the overall sampling and circuit-mining process using \textbf{gradient-based attribution}. \textbf{c)} shows the \textbf{mechanistic intervention} process. To improve TA robustness, we intervene in vulnerable attention modules via attention reweighting or zero-ablation. } 
    \label{fig:schematic}
\end{figure}

In this section, we introduce and analyze our training-free model interpretation and robustness process, as shown in \cref{fig:schematic}. Starting from the ViT architecture and the MHSA mechanism, we analyze the behavior of the concept vector sampling process, as visualized in \cref{fig:schematic}a, under the Linear Representation Hypothesis. The circuit mining and intervention process outlined in \cref{fig:schematic}b and \cref{fig:schematic}c is then established based on gradient attribution and the ViT architecture. 
\subsection{ViT Architecture}
To establish the foundation of our method, we briefly review the Multi-Head Self-Attention (MHSA) definition and the Residual Stream view of a Transformer~\cite{elhage2021mathematical}. Let the input to a specific layer of a ViT be $\mathbf{X} \in \mathbb{R}^{N \times d_{model}}$, where $N$ is the sequence length, and $d_{model}$ is the global residual stream dimension.
The MHSA block decomposes the representation into $H$ independent heads, projecting the data into a lower-dimensional subspace $d_{\text{head}} = d_{model} / H$. For a given head $h$, the computation is governed by the Query-Key (QK) routing circuit and the Output-Value (OV) feature circuit.
The QK circuit determines the pre-softmax attention logits $s_{i,j}$ from destination token $i$ to source token $j$:
\begin{equation}
s_{i,j} = \frac{(\mathbf{x}_i \mathbf{W}_Q^h)(\mathbf{x}_j \mathbf{W}_K^h)^T}{\sqrt{d_{\text{head}}}}
\end{equation}
where $\mathbf{x}_i, \mathbf{x}_j$ are row vectors from $\mathbf{X}$, and $\mathbf{W}_Q^h, \mathbf{W}_K^h, \mathbf{W}_V^h \in \mathbb{R}^{d_{model} \times d_{\text{head}}}$ are the learned projection matrices. The routing probabilities are obtained via softmax: $A_{i,j} = \frac{\exp(s_{i,j})}{\sum_k \exp(s_{i,k})}$.
Concurrently, the OV circuit extracts the local semantic feature vector at patch $j$: $\mathbf{v}_j = \mathbf{x}_j \mathbf{W}_V^h$. The head output vector written to destination token $i$ is
$\mathbf{o}_i = \sum\nolimits_{j=1}^{N} A_{i,j} \mathbf{v}_j$.
\subsection{Mechanistic Defense Task: Typographic Attack Robustness}
To motivate our interpretability framework, we formally define the task of mechanistic defense against typographic attacks in vision-language models like CLIP. A typographic attack occurs when a model is fooled by an adversarial text superimposed on an image (e.g., a cat labeled ``Goose'' being classified as a goose). Our goal is to localize and mitigate this vulnerability by decomposing the ViT as a collection of functional modules.
\subsubsection{ViT as Nested Functional Modules.}
We formalize the ViT $\mathcal{M}$ as a set $\mathcal{B}$ comprising a nested hierarchy of functional modules $\mathcal{B} = \{c_1, c_2, \dots, c_m\}$. These modules range from coarse structures (e.g., full residual blocks) to fine-grained components (e.g., attention modules and Multi-Layer Perceptrons~(MLPs)). 
Given a dataset distribution of clean input images $\mathbf{x} \sim \mathcal{D}$ with true semantic labels $y_{\text{img}}$, a task model $\mathcal{F}_{\mathcal{M}}$  based on $\mathcal{M}$, and a typographic perturbation function $\mathcal{T}(\mathbf{x}, y_{\text{text}})$ that superimposes the text distractor $y_{\text{text}}$ onto $\mathbf{x}$, the vulnerability is defined by the model prioritizing the text over the visual semantic content across the distribution:
\begin{equation}
\mathbb{P}_{\mathbf{x} \sim \mathcal{D}}\big(\mathcal{F}_{\mathcal{M}}(\mathcal{T}(\mathbf{x}, y_{\text{text}})) = y_{\text{text}}\big) \gg \mathbb{P}_{\mathbf{x} \sim \mathcal{D}}\big(\mathcal{F}_{\mathcal{M}}(\mathcal{T}(\mathbf{x}, y_{\text{text}})) = y_{\text{img}}\big)
\end{equation}
\subsubsection{Circuit Extraction and Intervention Objective.}
Mechanistic defense posits that this vulnerability is not distributed uniformly across the network but is localized within a specific ``typographic reading circuit.'' The defense task is to extract a minimal subset of computational modules $\mathcal{C}_{\text{lex}} \subset \mathcal{B}$ that route and propagate the adversarial lexical signal.
Let $\mathcal{I}$ be an intervention function applied specifically to the nodes in $\mathcal{C}_{\text{lex}}$ of ViT $\mathcal{M}$. Such interventions include activation ablation, attention reweighting, and others. The intervened model is denoted as $\mathcal{F}_{\mathcal{I}(\mathcal{M}, \mathcal{C}_{\text{lex}})}$. 
The mechanistic defense objective is to find the optimal subset $\mathcal{C}_{\text{lex}}^*$ that maximizes robustness against the attack while minimizing the degradation of general capabilities on clean data, constrained by the sparsity of the localized circuit:
\begin{equation}
\mathcal{C}_{\text{lex}}^* = \underset{\mathcal{C}_{\text{lex}} \subset \mathcal{B}}{\arg\min} |\mathcal{C}_{\text{lex}}| \quad \text{s.t.} \quad 
\begin{cases}
\mathbb{E}_{\mathbf{x} \sim \mathcal{D}} \left[ \mathcal{L} \left( \mathcal{F}_{\mathcal{I}(\mathcal{M},\mathcal{C}_{\text{lex}})}(\mathcal{T}(\mathbf{x}, y_{\text{text}})), y_{\text{img}} \right) \right] < \epsilon_{robust} \\
\mathbb{E}_{\mathbf{x} \sim \mathcal{D}} \left[ \mathcal{L} \left( \mathcal{F}_{\mathcal{I}(\mathcal{M}, \mathcal{C}_{\text{lex}})}(\mathbf{x}), y_{\text{img}} \right) \right] < \epsilon_{benign}
\end{cases}
\end{equation}
where $\mathcal{L}$ is the utility loss, e.g., Classification Error, $\epsilon_{robust}$ guarantees TA robustness, and $\epsilon_{benign}$ ensures the preservation of the model's core vision capabilities. To efficiently discover the members of $\mathcal{C}_{\text{lex}}^*$, we must systematically trace the attribution of the $y_{\text{text}}$ signal back through the computational graph.
\subsection{Information Flow Attribution}
\label{sec:grad-attr}
To trace the adversarial routing circuit, we attribute the flow of specific semantic concepts to the routing decisions made by the QK circuit. We probe the head's $d_{\text{head}}$-dimensional output space with a direction vector $\mathbf{u} \in \mathbb{R}^{d_{\text{head}}}$. 
We define the projected concept magnitude at source patch $j$ as $V_j(\mathbf{u}) = \langle \mathbf{v}_j, \mathbf{u} \rangle$, and the concept strength aggregated at destination token $i$ as:
\begin{equation}
F_i(\mathbf{u}) = \langle \mathbf{o}_i, \mathbf{u} \rangle = \sum_{j=1}^{N} A_{i,j} V_j(\mathbf{u})
\end{equation}
We formalize attribution as the partial derivative of the concept strength with respect to the pre-softmax logit $s_{i,j}$. Applying the chain rule through the softmax function yields the following attribution map:
\begin{align}
\frac{\partial F_i(\mathbf{u})}{\partial s_{i,j}} &= \sum_k \frac{\partial A_{i,k}}{\partial s_{i,j}} V_k(\mathbf{u}) \nonumber \\
&= A_{i,j}(1 - A_{i,j})V_j(\mathbf{u}) - \sum_{k \neq j} A_{i,j} A_{i,k} V_k(\mathbf{u}) \nonumber \\
&= A_{i,j} \big( V_j(\mathbf{u}) - F_i(\mathbf{u}) \big)
\label{eq:grad}
\end{align}
\noindent \textbf{Interpretation.} The gradient attribution of a patch's attention logit to the pseudo-concept vector $\mathbf{u}$ can be transformed into the product of the routing gate ($A_{i,j}$) and the marginal utility of the patch ($V_j(\mathbf{u}) - F_i(\mathbf{u})$). This formulation jointly considers the attention activation and concept alignment to ensure faithful attribution. 
\subsection{The Stochastic Lottery and Polysemantic Interference}
A core difficulty in the efficient application of concept-based interpretation to deep neural networks is the acquisition of the concept vector $\mathbf{u}$ without computationally expensive and data-intensive dictionary learning. To relax this requirement and promote training-free model interpretability, we apply stochastic sampling to mechanistic interpretability research. Based on the Linear Representation Hypothesis~\cite{park2024linear, elhage2022superposition} that the dense hidden space representation is a superposition of several distinct concept vectors, we propose to sample and evaluate random hidden space vectors for interpreting neural network behaviors. This formulation intuitively connects to the Lottery Ticket Hypothesis~\cite{frankle2018the}: By considering random vector samples as the linear probe initialization and the lexical information as the learning target, there exists a sparse subset of the model, in this case, a subset of sampled vectors, that preserves the model capacity to achieve the learning target.
The primary mathematical threat to this stochastic lottery is polysemantic interference. For the analysis, we generate $K$ independent random vectors per head, denoting each as $\mathbf{u} \sim \mathcal{N}(\mathbf{0}, \frac{1}{d_{\text{head}}}\mathbf{I})$. Let us isolate a target concept $\mathbf{c}_{\text{target}} \in \mathbb{R}^{d_{\text{head}}}$ ($\|\mathbf{c}_{\text{target}}\| = 1$). We orthogonally decompose a patch $i$'s value vector $\mathbf{v}_i$:
\begin{equation}
\mathbf{v}_i = \alpha_i \mathbf{c}_{\text{target}} + \boldsymbol{\xi}_i,
\end{equation}
where $\alpha_i \in \mathbb{R}$ is the true concept strength and $\boldsymbol{\xi}_i \in \mathbb{R}^{d_{\text{head}}}$ is the polysemantic interference vector, such that $\langle \mathbf{c}_{\text{target}}, \boldsymbol{\xi}_i \rangle = 0$.
When projecting onto a random feature direction $\mathbf{u}$, the activation is $\langle \mathbf{v}_i, \mathbf{u} \rangle$:
\begin{equation}
Z_i \propto \langle \mathbf{v}_i, \mathbf{u} \rangle = \underbrace{\alpha_i \langle \mathbf{c}_{\text{target}}, \mathbf{u} \rangle}_{\text{Signal } S_{\mathbf{u},i}} + \underbrace{\langle \boldsymbol{\xi}_i, \mathbf{u} \rangle}_{\text{Interference } I_{\mathbf{u},i}},
\end{equation}
where $S_{\mathbf{u},i}$ is the true concept signal and $I_{\mathbf{u},i}$ denotes the polysemantic interference.
\subsubsection{Bounding Interference via Concentration of Measure.}
By standard Gaussian tail bounds, the maximum interference across all $N$ patches is bounded. Denoting $\|\boldsymbol{\xi}_{\text{max}}\| = \max_{i \in \{1\dots N\}} \|\boldsymbol{\xi}_i\|$ as the
largest interference-vector norm across patches, the maximum error in estimating the true concept strength is bounded with high probability:
\begin{equation}
\label{eq:condition}
\max_{i \in \{1 \dots N\}} |I_{\mathbf{u},i}| \le \|\boldsymbol{\xi}_{\text{max}}\| \sqrt{\frac{4 \log N}{d_{\text{head}}}}
\end{equation}
For a clean attribution map, the signal of the weakest on-concept patch $k= \arg\min_{i \in \text{on}} \alpha_i$ must dominate the maximum interference by a margin $\tau$:
\begin{equation}
\label{eq:separation}
S_{\mathbf{u},k} > \max_i |I_{\mathbf{u},i}| + \tau .
\end{equation}
The probability that a single random probe satisfies \cref{eq:separation} is
\begin{equation}
\label{eq:psingle}
p \approx \exp\!\left(
  - \frac{\left( \|\boldsymbol{\xi}_{\text{max}}\| \sqrt{4 \log N} + \tau \sqrt{d_{\text{head}}} \right)^2}{2(\alpha_k)^2}
\right).
\end{equation}
Since every on-concept patch carries the concept at least as strongly as $\alpha_k$, a probe satisfying \cref{eq:separation} resolves all of them simultaneously, yielding a concept-faithful attribution map. By construction of the text-injected inputs, the injected region carries the lexical concept strongly, so $\alpha_k$ is bounded well above zero. We defer the detailed derivation to the Appendix.
\subsubsection{The Advantage of the MHSA Bottleneck.}
Considering a sample size of $K$, we define success as the event that at least one sampled $\mathbf{u}$ satisfies \cref{eq:separation} with respect to $\mathbf{c}_{\text{target}}$. Then, to achieve a higher probability of finding clean feature maps ($P_{\text{success}}$), the sample-and-interpret process requires $K$ lottery tickets per head:
\begin{equation}
K \ge \frac{\log(1 - P_{\text{success}})}{\log(1 - p)}
\end{equation}
This inequality justifies operating in $d_{\text{head}}$ rather than $d_{\text{model}} \gg d_{\text{head}}$. In the global stream, the interference norm $\|\boldsymbol{\xi}_{\text{max}}\|^2$ contains the sum of all features across all heads, causing the required $K$ to grow rapidly. Isolating the attention-head-specific subspace pushes head-excluded concepts into the null space, which significantly reduces $\|\boldsymbol{\xi}_{\text{max}}\|^2$, increases $p$, and allows a computationally feasible $K$ to reliably obtain concept basis vectors as winning tickets, yielding highly interpretable semantic vectors without parameter training.
\subsection{Mechanistic Intervention}
Given extracted lexical circuits consisting of attention heads with high text attribution, we perform attention reweighting on each attention head to simultaneously maintain representation consistency while reducing typographic information in the object classification task, and apply zero ablation to remove lexical distraction from all patch tokens under VQA evaluation, as shown in \cref{fig:schematic}c.
\subsubsection{Attention Reweighting.}
Following Dyslexify~\cite{dyslexify}, for input sequence of length $N$, given the original attention map for the <cls> token $\mathbf{a}_{\text{<cls>}} = [A_{0,0}, A_{0,1}, A_{0,2}, \allowbreak \ldots A_{0, N-1}] \in \mathbb{R}^{N}$ and a control parameter $a\in [0,1]$, the reweighted attention map $\mathbf{a}_{\text{<cls>}}' \in \mathbb{R}^{N}$ is 
\begin{equation}
\label{eq:reweight}
\mathbf{a}'_{\text{<cls>}}[i] = a \cdot \mathbb{I}(i=0) + \mathbf{a}_{\text{<cls>}}[i] \frac{1-a}{\sum_{j=1}^{N-1} \mathbf{a}_{\text{<cls>}}[j]} \cdot \mathbb{I}(i \neq 0)
\end{equation}
where the  $i$ and $j$ denote the token location in the attention maps, and the token order follows the ViT definition~\cite{dosovitskiy2021an}  with the first token being the <cls> token and the following being patch tokens. 
The reweighted attention map effectively reduces the information passed from textual patch tokens in a text-focused attention head, thereby suppressing the lexical circuit. Throughout our experiments, we set $a=1$ to achieve maximal intervention.
\subsubsection{Zero Ablation.}
Attention Reweighting does not apply to LVLMs, which often do not implement and train a dedicated class token and instead rely on all patch token embeddings for visual information processing. Therefore, to evaluate the quality and faithfulness of the extracted lexical circuits, we use simple zero ablation, in which a zero vector replaces the output of the vulnerable modules. 

%% file: sec/4_experiments.tex
\section{Experiments}
To validate the efficacy of our method in extracting faithful typographic circuits and improving visual perception under lexical interference, we conduct object classification experiments using five ViT CLIP models of varying sizes pretrained on LAION-2B~\cite{schuhmann2022laion} and VQA experiments with several popular LVLMs.
\subsection{Experimental Settings}
\subsubsection{Evaluation Datasets.}
Following prior works~\cite{denseprefix, dyslexify}, we use RTA-100~\cite{denseprefix}, Disentangling~\cite{disentangle}, and PAINT~\cite{ilharco2022patching} datasets for the object classification experiments. In addition, we construct a new dataset IN-100-Text from ImageNet-100~\cite{clane_im100} by adding realistic, contextually coherent text distractors using Qwen-Image-Edit~\cite{wu2025qwen}. For each image, we sample a label from the IN-100 class vocabulary to create a semantic conflict between the visual content and the lexical distractor, and render the label in one of seven diverse visual styles detailed in the Appendix. The pixel-level edit area with a max-channel difference greater than 20/255 has a mean/median of 16.3\%/11.5\%. For VQA evaluation, we use RIO-Bench~\cite{waseda2025read}, a large VQA dataset consisting of TA images. We specifically evaluate our method on the attacked multiple-choice VQA, comprising easy, medium, and hard subsets. \looseness=-1
\subsubsection{Baseline Methods.}
We demonstrate the advantage of our proposed method in improving TA robustness and enhancing model performance under lexical interference by comparing to Defense-Prefix~\cite{denseprefix} and Dyslexify~\cite{dyslexify} for object classification. Defense-Prefix Training is a supervised method that learns a defense-prefix token embedding. Dyslexify~\cite{dyslexify} is a training-free method based on attention statistics and greedy circuit mining. To further demonstrate the efficacy of our method, we apply it to several LVLMs and report performance with and without our intervention. 
\subsubsection{Implementation Details.}
For object classification experiments, we test our method on ViT-B/16, ViT-L/14, ViT-H/14, ViT-g/14, and ViT-bigG/14 models pretrained on the LAION-2B~\cite{schuhmann2022laion} dataset. Based on prior observation that concept emerges in the late stages of ViTs~\cite{gandelsman2024interpreting}, we systematically examine the last $20\%$ of the transformer blocks in the ViTs. For each attention module, we sample 16 times the hidden dimension of random concept vectors and perform gradient-based attribution as described in \cref{sec:grad-attr}.  We use the ImageNet-1K~\cite{imnet} training set as the base dataset for text injection augmentation. Only $0.1\%$ of the training set is used in each run. 
The one-time extraction runs on 1,280 images and takes \textbf{under one minute on a single A100} across all five backbones (ViT-B/16: 7.5\,s; ViT-L/14: 14.0\,s; ViT-H/14: 24.0\,s; ViT-g/14: 32.1\,s; ViT-bigG/14: 45.7\,s). At test-time, the intervention incurs \textbf{near-zero overhead} because it operates on a fixed set of head indices.
\subsection{Vulnerability Identification}
Following~\cref{eq:grad}, we consider in a practical setup with an input image $I$ with known text location $M \subseteq \{0, \ldots N - 1\}$, the text-corresponding patch locations yield the text mask $\mathbf{m}^+ \in \mathbb{R}^{N}$ with $\mathbf{m}[p]=\mathds{1}_M(p)$, where $p$ is the token index and $\mathds{1}_M$ is the indicator function given the set $M$. Then the inverse text mask $\mathbf{m}^- =  \mathbf{1}_{N} - \mathbf{m}^+$, with the exception that the class token mask value is always zero. The Mask Attribution Score~(MAS) and normalized Text Attribution Score~(nTAS) with respect to the destination token index $i$ are defined as:
\begin{align}
\label{eq:tas}
    \text{MAS}_i(\mathbf{u}, \mathbf{m}) &= \frac{1}{\langle\mathbf{1}_N, \mathbf{m}\rangle} \sum_{j=1}^N \text{ReLU}(\frac{\partial F_i(\mathbf{u})}{\partial s_{i,j}}) \cdot \mathbf{m}[j],  \\
    \text{nTAS}_i(\mathbf{u}, \mathbf{m}^+, \mathbf{m}^-) &= \frac{\text{MAS}_i(\mathbf{u}, \mathbf{m}^+) }{\text{MAS}_i(\mathbf{u}, \mathbf{m}^+)  + \text{MAS}_i(\mathbf{u}, \mathbf{m}^-) }.
\end{align}
where $\mathbf{m}$ is a vector mask and $\mathbf{m}[\cdot]$ indexes the mask.  
In \cref{eq:tas}, the MAS considers only positive gradient attribution to the patch token sequence, consistent with the additive nature of the Linear Representation Hypothesis~\cite{park2024linear, elhage2022superposition}. Normalized TAS jointly accounts for object and lexical focus and scales the score to the $[0, 1]$ range. We take the mean of all sample vector $\mathbf{u}$ scores over an augmented unlabeled dataset as the score of the tested module. Attention head modules within a layer are compared with a simple z-test. The z-score threshold is either selected with a calibration dataset or defaults to one. This choice is not critical near the default: on ViT-H/14 (RTA-100), $z \in \{0.5, 1.0, 2.0\}$ selects 24/15/4 heads, yielding Object Classification Accuracy (OCA) = 75.9/76.3/66.5\% and Text Confusion Rate (TCR) = 14.0/15.4/27.1\%. The clean ImageNet-100 classification accuracy remains at 83.2/83.3/83.9\%. Robustness and clean accuracy are stable around z = 1, and only at z = 2 do too few heads remain in the lexical circuit.
\subsection{Improved TA Robustness}
In \cref{tab:defense}, we demonstrate the efficacy of our method in improving the TA robustness of various CLIP ViT models with two metrics: object classification accuracy and text confusion rate. Object classification accuracy measures the classification accuracy on the target object depicted in the input image. Text confusion rate measures the accuracy of the models in predicting the text label of the text distractor, and a lower confusion rate indicates a better suppression of the lexical confusion signal. As reported in \cref{tab:defense}, our method improves the object classification accuracy of all five tested CLIP ViT models from the base model to the bigG variant. The object classification accuracy increases significantly, accompanied by a large decline in the text confusion rate. With minimal data and computational requirements, our method remains competitive with other methods. Notably, our method incurs an accuracy trade-off of less than $1\%$ in exchange for a significant gain in robustness, as shown in the Appendix.
\begin{table}[t]
    \centering
    \setlength{\tabcolsep}{4pt}
     \caption{Robustness improvement across ViT backbones on Zero-shot image classification. \textbf{OCA}: Object Classification Accuracy ($\uparrow$), \textbf{TCR}: Text Confusion Rate ($\downarrow$). The \textcolor{green!60!black}{green colored numbers} indicate the improvement in OCA, and the \textcolor{blue}{blue colored numbers} indicate a decrease in TCR. \textbf{w. Int.} indicate \textit{with intervention}.}
     \label{tab:defense}
    \resizebox{1\textwidth}{!}{%
    \begin{tabular}{ll cc cc cc cc}
        \toprule
        \multirow{2}{*}{\textbf{Model}} & \multirow{2}{*}{\textbf{Method}} & \multicolumn{2}{c}{\textbf{RTA-100}} & \multicolumn{2}{c}{\textbf{Disentangling}} & \multicolumn{2}{c}{\textbf{PAINT}} & \multicolumn{2}{c}{\textbf{IN-100-Text}}  \\
        \cmidrule(lr){3-4} \cmidrule(lr){5-6} \cmidrule(lr){7-8} \cmidrule(lr){9-10}
         & & \textbf{OCA}($\uparrow$) & \textbf{TCR}($\downarrow$) & \textbf{OCA}($\uparrow$) & \textbf{TCR}($\downarrow$) & \textbf{OCA}($\uparrow$) & \textbf{TCR}($\downarrow$) & \textbf{OCA}($\uparrow$)& \textbf{TCR}($\downarrow$) \\
        % ViT-B-16 Data
\midrule
\midrule
\multirow{2}{*}{ViT-B/16}
& Baseline               & 56.3 & 30.8 & 52.2 & 44.4 & 60.2 & 33.0 & 54.6 & 33.0 \\
& w. Int.     & 68.7 \textcolor{green!60!black}{(+12.4)} & 12.6 \textcolor{blue}{(-18.2)} & 88.3 \textcolor{green!60!black}{(+36.1)} & 11.7 \textcolor{blue}{(-32.7)} & 73.8 \textcolor{green!60!black}{(+13.6)} & 16.5 \textcolor{blue}{(-16.5)} & 74.2 \textcolor{green!60!black}{(+19.6)} & 7.1 \textcolor{blue}{(-25.9)} \\
\midrule
\multirow{2}{*}{ViT-L/14}
& Baseline               & 54.6 & 39.0 & 51.7 & 47.8 & 61.2 & 33.0 & 58.2 & 32.9 \\
& w. Int.     & 68.9 \textcolor{green!60!black}{(+14.3)} & 21.2 \textcolor{blue}{(-17.8)} & 68.3 \textcolor{green!60!black}{(+16.6)} & 31.1 \textcolor{blue}{(-16.7)} & 68.9 \textcolor{green!60!black}{(+7.7)} & 22.3 \textcolor{blue}{(-10.7)} & 74.9 \textcolor{green!60!black}{(+16.7)} & 12.1 \textcolor{blue}{(-20.8)} \\
\midrule
\multirow{2}{*}{ViT-H/14}
& Baseline               & 53.4 & 42.0 & 46.1 & 53.3 & 49.5 & 46.6 & 56.6 & 36.9 \\
& w. Int.     & 76.2 \textcolor{green!60!black}{(+22.8)} & 14.4 \textcolor{blue}{(-27.6)} & 82.2 \textcolor{green!60!black}{(+36.1)} & 17.2 \textcolor{blue}{(-36.1)} & 75.7 \textcolor{green!60!black}{(+26.2)} & 14.6 \textcolor{blue}{(-32.0)} & 79.1 \textcolor{green!60!black}{(+22.5)} & 9.5 \textcolor{blue}{(-27.4)} \\
\midrule
\multirow{2}{*}{ViT-g/14}
& Baseline               & 50.3 & 45.8 & 58.3 & 41.1 & 53.4 & 39.8 & 57.0 & 36.4 \\
& w. Int.     & 68.8 \textcolor{green!60!black}{(+18.5)} & 23.4 \textcolor{blue}{(-22.4)} & 81.7 \textcolor{green!60!black}{(+23.4)} & 17.8 \textcolor{blue}{(-23.3)} & 75.7 \textcolor{green!60!black}{(+22.3)} & 18.4 \textcolor{blue}{(-21.4)} & 76.4 \textcolor{green!60!black}{(+19.4)} & 12.7 \textcolor{blue}{(-23.7)} \\
\midrule
\multirow{2}{*}{ViT-bigG/14}
& Baseline               & 61.0 & 32.5 & 48.3 & 51.1 & 49.5 & 38.8 & 62.3 & 31.0 \\
& w. Int.     & 75.7 \textcolor{green!60!black}{(+14.7)} & 15.3 \textcolor{blue}{(-17.2)} & 72.8 \textcolor{green!60!black}{(+24.5)} & 26.7 \textcolor{blue}{(-24.4)} & 79.6 \textcolor{green!60!black}{(+30.1)} & 10.7 \textcolor{blue}{(-28.1)} & 80.6 \textcolor{green!60!black}{(+18.3)} & 8.9 \textcolor{blue}{(-22.1)} \\
        \bottomrule
    \end{tabular}%
    }
\end{table}
\subsection{Object Classification Evaluation}
To make a fair comparison with Dyslexify~\cite{dyslexify} and to fit the training-free test setup, we use the same $0.1\%$ fraction of the ImageNet-1K training set as the base dataset to compute attention scores for its greedy circuit mining. As shown in \cref{tab:compare}, our method improves object classification accuracy across CLIP ViT variants and achieves a significant boost in average robustness over the supervised baseline. With concept-based mechanistic interpretability, our method achieves higher average robustness than the prior method~\cite{dyslexify}. Our joint modeling of the attention mechanism and concept distribution for circuit mining effectively isolates TA vulnerability in the transformer. This enables simple statistical vulnerability detection that requires minimal labeled data and provides greater TA robustness (an average OCA improvement of 6.1\%) than a greedy search guided by large labeled image sets on noisy attention maps. We report detailed trade-offs on clean image perception in the Appendix. Our method incurs minimal accuracy trade-offs, consistent with other methods. 
\begin{table}[t]
    \centering
     \caption{\textbf{Comparison across methods against typographic attacks.} Numbers are zero-shot classification accuracy in percentage. * indicates results reported by Dyslexify~\cite{dyslexify}, which requires iterative evaluation on full ImageNet-100 training set.}
    \label{tab:compare}
    \resizebox{0.8\linewidth}{!}{
    \begin{tabular}{c|c|c|c|c|c|c}
    \toprule
    \textbf{Method} & \textbf{Model} & \textbf{RTA-100} & \textbf{Disentangling} & \textbf{PAINT} & \textbf{IN-100-Text} & \textbf{IN-100} \\
    \midrule
    \midrule
    \multirow{6}{*}{Defense-Prefix~\cite{denseprefix}} 
    & ViT-B/16 & 65.6 & 84.4 & 68.0 & 69.9 & \textit{75.4}\\
    & ViT-L/14 & 62.9 & \textit{77.2} & 71.8 & 71.6  & \textit{79.8}\\
    & ViT-H/14 & 63.5 & 67.2 & 64.1 & 70.4 &\textit{ 83.4} \\
    & ViT-g/14 & 58.1 & 60.0 & 66.0 & 66.7 & 83.2\\
    & ViT-bigG/14 & 67.8 & 50.0 & 68.9 & 71.7 & \textit{85.4}\\
    \cmidrule{2-7}
    & \textit{Average} & 63.6 & 67.8 & 67.8 & 70.1 & \textbf{81.4}\\
    \midrule
    \multirow{6}{*}{Dyslexify~\cite{dyslexify}} 
    & ViT-B/16 & 67.9 & 83.9 & 69.9 & 73.5 & 75.3\\ 
    & ViT-L/14 & 66.6 & 68.9 & \textit{72.8} & 74.4 & 79.5\\ 
    & ViT-H/14 & 66.1 & 53.9 & 68.9 & 70.3 & \textit{83.4}\\ 
    & ViT-g/14 & 67.0 & 75.6 & \textit{77.7} & 75.0 & 83.0\\ 
    & ViT-bigG/14 & 67.5 & 55.6 & 61.2 & 70.6 & 85.0 \\ 
    \cmidrule{2-7}
    & \textit{Average} & 67.0 & 67.6 & 70.1 & 72.8 & 81.3 \\
    \midrule
    \multirow{6}{*}{Dyslexify~\cite{dyslexify}*} 
    & ViT-B/16 & 68.3 & 85.0 & 72.7 & - & 75.0\\ 
    & ViT-L/14 & 71.0 & 60.6 & 76.4 & - & 79.5 \\ 
    & ViT-H/14 & 68.3 & 72.2 & 70.9 & - & \textit{83.4}\\ 
    & ViT-g/14 & 62.0 & 67.2 & 71.8 & - & 82.6 \\ 
    & ViT-bigG/14 & 72.9 & 68.3 & 69.1 & -& 84.7 \\ 
    \cmidrule{2-7}
    & \textit{Average} & 68.5 & 70.7 & 72.2 & - & 81.0\\
    \midrule
    \multirow{6}{*}{Ours~(nTAS)} 
    & ViT-B/16 & \textit{68.7} & \textit{88.3} & \textit{73.8} & \textit{74.2} & 74.2\\
    & ViT-L/14 & \textit{68.9} & 68.3 & 68.9 & \textit{74.9} &  \textit{79.8}\\
    & ViT-H/14 & \textit{76.2} & \textit{82.2} & \textit{75.7} & \textit{79.1} & 82.9 \\
    & ViT-g/14 & \textit{68.8} & \textit{81.7} & 75.7 & \textit{76.4} & \textit{83.3} \\
    & ViT-bigG/14 & \textit{75.7} & \textit{72.8} & \textit{79.6} & \textit{80.6} & 84.6 \\
    \cmidrule{2-7}
    & \textit{Average} & \textbf{71.7} & \textbf{78.7} & \textbf{74.7} & \textbf{77.0} & 81.0 \\
    \bottomrule
    \end{tabular}
    }
\end{table}
\subsection{Application on LVLMs}
In order to further demonstrate the efficacy and application of our method, we apply the circuit mining pipeline to the vision encoder of several latest LVLMs that follow the standard ViT-MLP-LLM architecture: Qwen3-VL~\cite{bai2025qwen3}, InternVL3.5~\cite{wang2025internvl3}, and Gemma3~\cite{Gemma_Team2025-wf}. Given that many LVLMs do not implement a <cls> token, we perform attribution based on the first visual token. As ViTs are known to repurpose redundant patch tokens to process global information~\cite{darcet2024vision}, the first visual token, fixed at the top-left corner, is highly likely to serve as an emergent <cls> surrogate. 
We compare our method with a vanilla LVLM without intervention. In \cref{tab:LVLM}, we show the VQA accuracy on RIO-Bench \textit{obj-attack} split, which requires the models to ignore the text literal in the image and answer questions based on the visual cues. Our method improves the VQA accuracy across Qwen3-VL and Gemma3 model variants. The tradeoff on clean image VQA is further tested on RIO-Bench \textit{clean} split. Our method incurs a minimal $-0.3\%$ to $+0.6\%$ effect on clean image VQA, with detailed numbers in the Appendix.\looseness=-1
\begin{table}[t]
\centering
\caption{VQA accuracy on RIO-Bench splits. $\mathrm{\Delta}$ represents the improvement ($\text{Ours} - \text{Base}$). Improvements $> 0.2$ are highlighted in bold.}
\label{tab:LVLM}
\resizebox{1\textwidth}{!}{% 
\setlength{\tabcolsep}{5pt}
\begin{tabular}{lcccccccccccc}
\toprule
\multirow{2}{*}{\textbf{Model}} & \multicolumn{3}{c}{\textbf{Easy}} & \multicolumn{3}{c}{\textbf{Medium}} & \multicolumn{3}{c}{\textbf{Hard}} & \multicolumn{3}{c}{\textbf{Overall Average}} \\
\cmidrule(lr){2-4} \cmidrule(lr){5-7} \cmidrule(lr){8-10} \cmidrule(lr){11-13}
 & Ours & Base & $\mathrm{\Delta}\uparrow$ & Ours & Base & $\mathrm{\Delta}\uparrow$ & Ours & Base & $\mathrm{\Delta}\uparrow$ & Ours & Base & $\mathrm{\Delta}\uparrow$ \\
\midrule
Qwen3-VL-4B & 69.71 & 68.29 & \textbf{1.42} & 66.52 & 65.67 & \textbf{0.85} & 55.53 & 54.90 & \textbf{0.63} & 63.92 & 62.95 & \textbf{0.97} \\
Qwen3-VL-8B & 75.81 & 74.19 & \textbf{1.62} & 73.34 & 71.23 & \textbf{2.11} & 65.92 & 64.91 & \textbf{1.01} & 71.69 & 70.11 & \textbf{1.58} \\
Qwen3-VL-30B-A3B & 69.11 & 67.34 & \textbf{1.77} & 66.87 & 66.71 & 0.16 & 62.35 & 60.90 & \textbf{1.45} & 66.11 & 64.98 & \textbf{1.13} \\
\midrule
InternVL3.5-8B & 63.96 & 63.87 & 0.09 & 61.18 & 60.77 & \textbf{0.41} & 50.28 & 50.09 & 0.19 & 58.47 & 58.24 & \textbf{0.23} \\
InternVL3.5-14B & 57.01 & 56.95 & 0.06 & 55.37 & 55.65 & -0.28 & 46.87 & 46.62 & \textbf{0.25} & 53.08 & 53.07 & 0.01 \\
\midrule
Gemma3-4B & 49.72 & 47.98 & \textbf{1.74} & 47.98 & 46.46 & \textbf{1.52} & 40.52 & 38.60 & \textbf{1.92} & 46.07 & 44.35 & \textbf{1.73} \\
Gemma3-12B & 52.27 & 49.91 & \textbf{2.36} & 49.84 & 48.74 & \textbf{1.10} & 47.22 & 46.08 & \textbf{1.14} & 49.78 & 48.24 & \textbf{1.53} \\
\bottomrule
\end{tabular}
}
\end{table}

%% file: sec/5_discussion.tex
\section{Discussion and Analysis}
\subsection{Qualitative Study}
\begin{figure}[t]
    \centering
    \begin{tabular}{c | c}
    \includegraphics[width=0.495\linewidth,trim={0 0 0 300pt},clip]{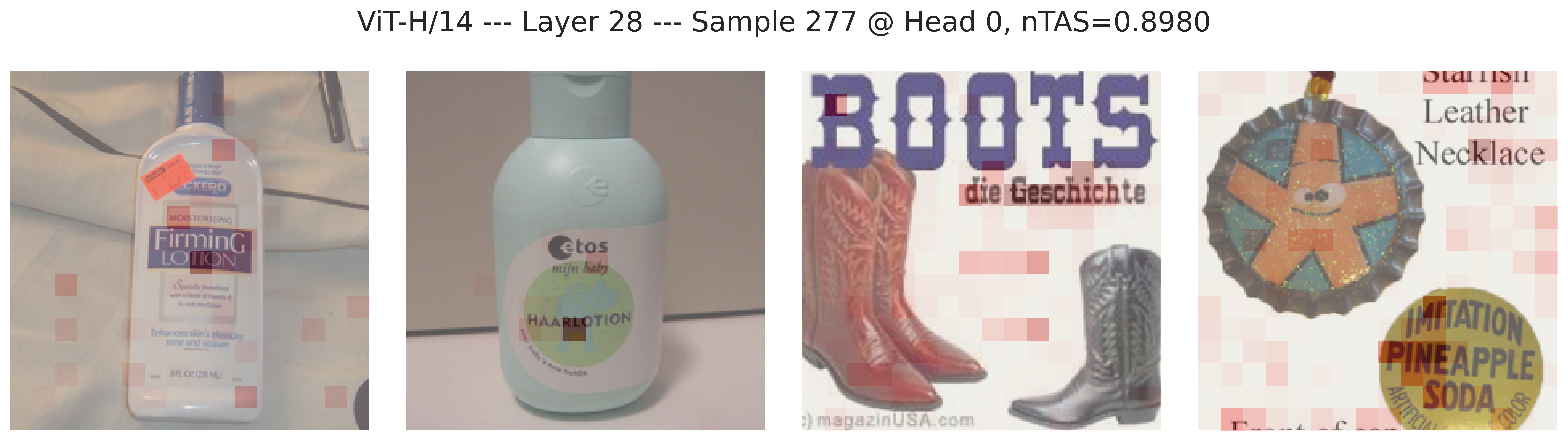} &
    \includegraphics[width=0.495\linewidth,trim={0 0 0 300pt},clip]{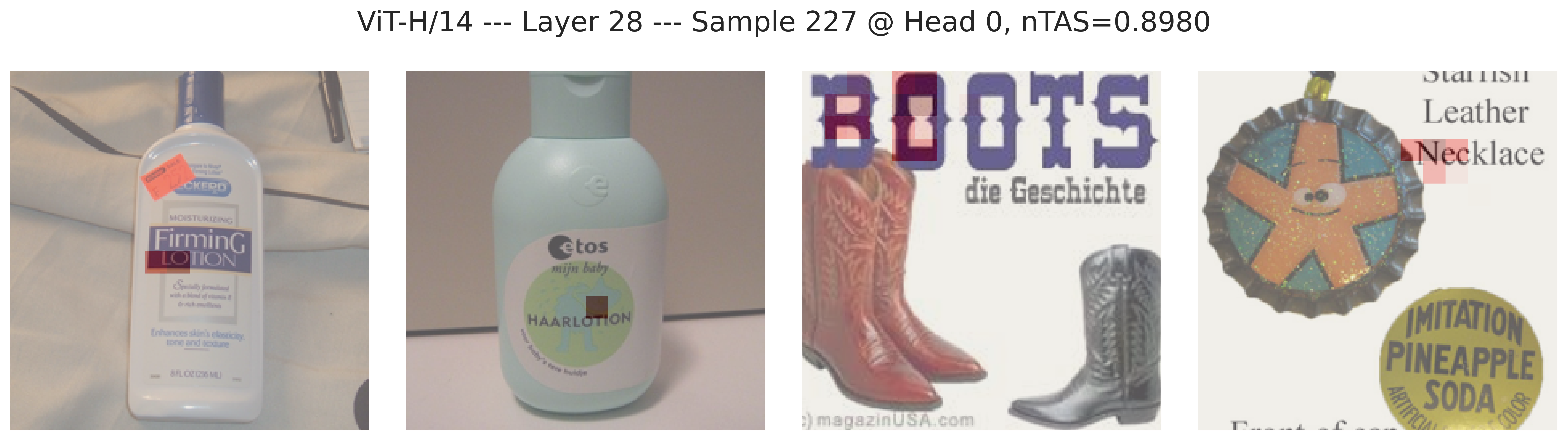}
    \end{tabular}
    \caption{\textbf{Attribution Map of Aligned and Unaligned Concept Vectors.} \textbf{Left:} Image patches are attributed with an unaligned concept vector. The resulting attribution map does not create an interpretable region. \textbf{Right:} The image patches are attributed with an aligned vector that produces the nTAS score of 0.8980. The resulting attribution map is highly concentrated in the lexical shapes in the images. }
    \label{fig:alignment}
\end{figure}

\begin{figure}[t]
    \begin{center}
    \resizebox{1\linewidth}{!}{
    \begin{tabular}{c | c}
        \includegraphics[width=0.495\linewidth]{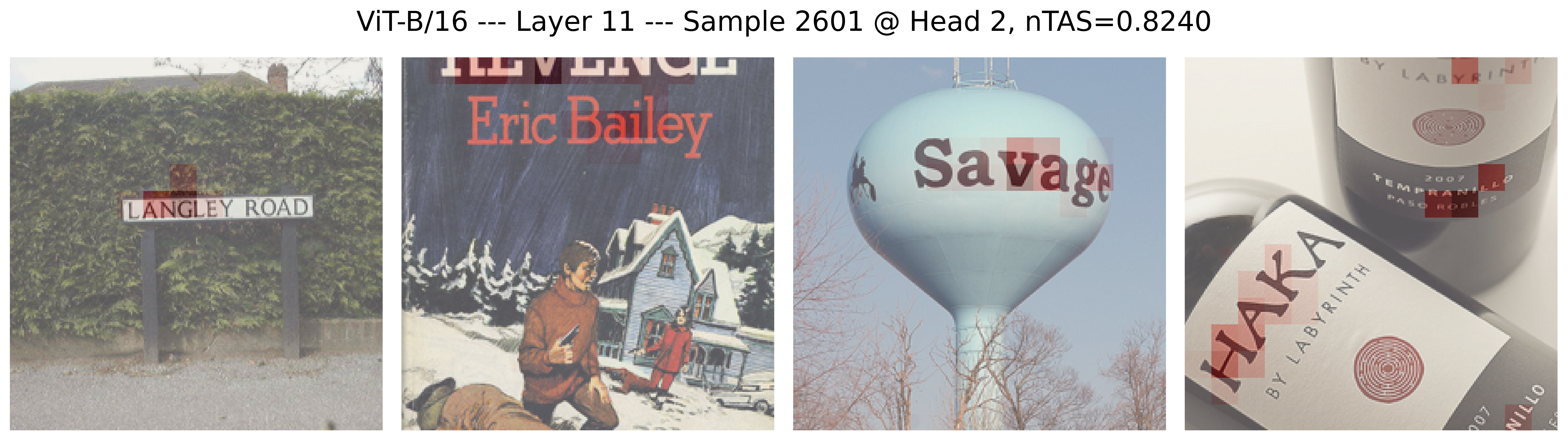} & 
        \includegraphics[width=0.495\linewidth]{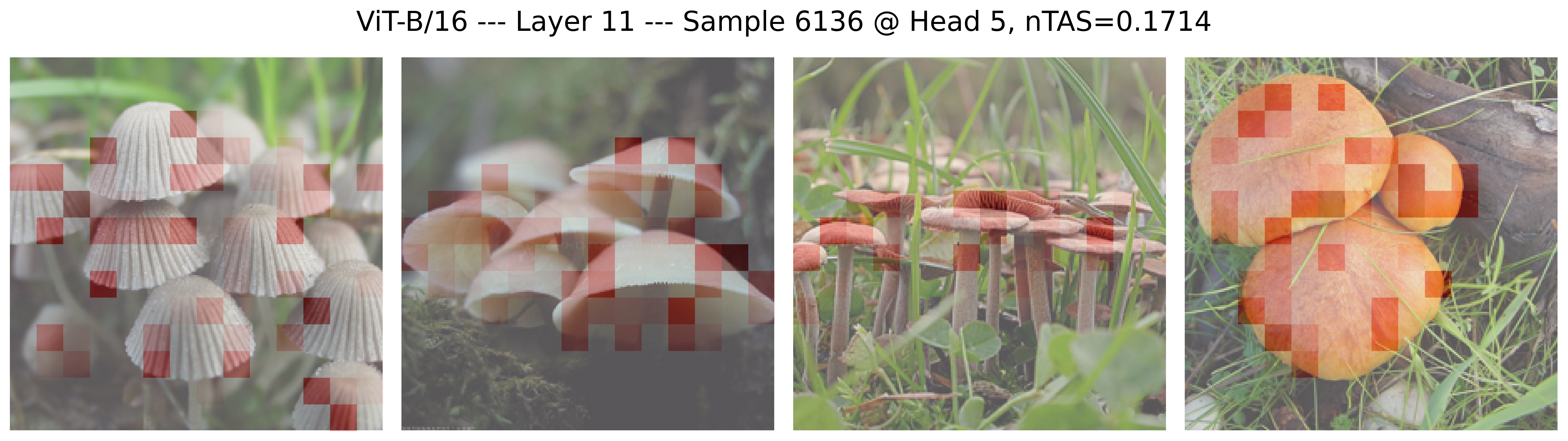} \\
        \includegraphics[width=0.495\linewidth]{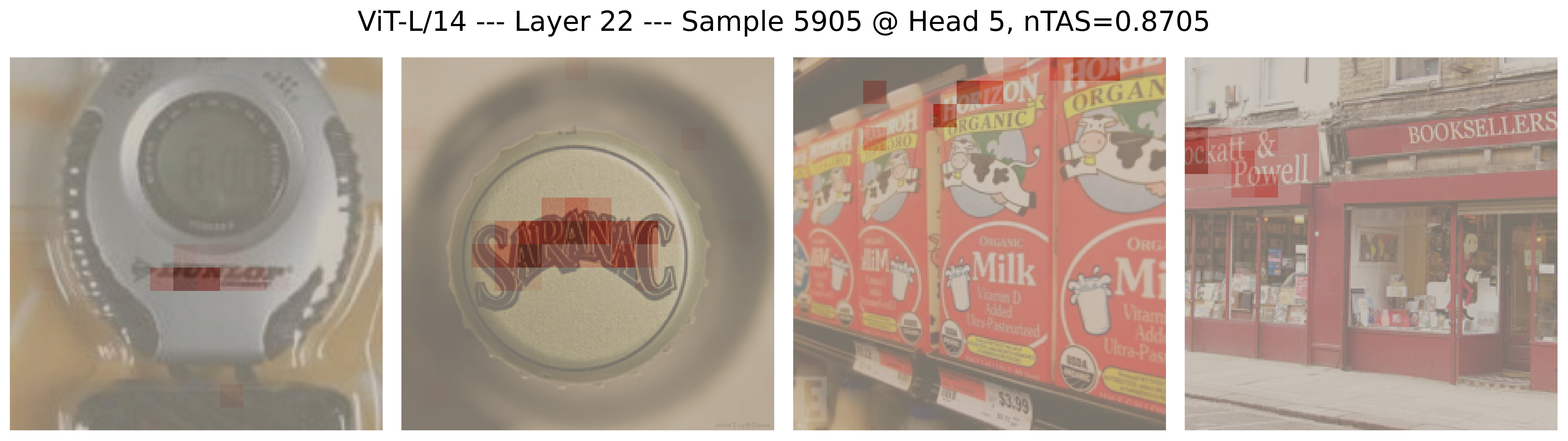} &
        \includegraphics[width=0.495\linewidth]{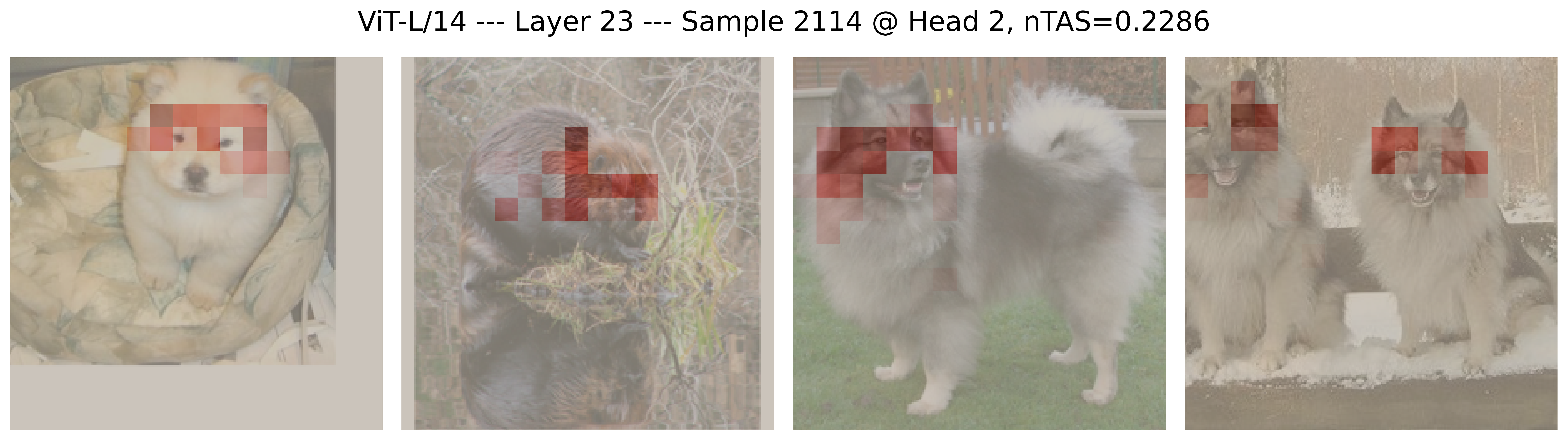} \\
        \includegraphics[width=0.495\linewidth]{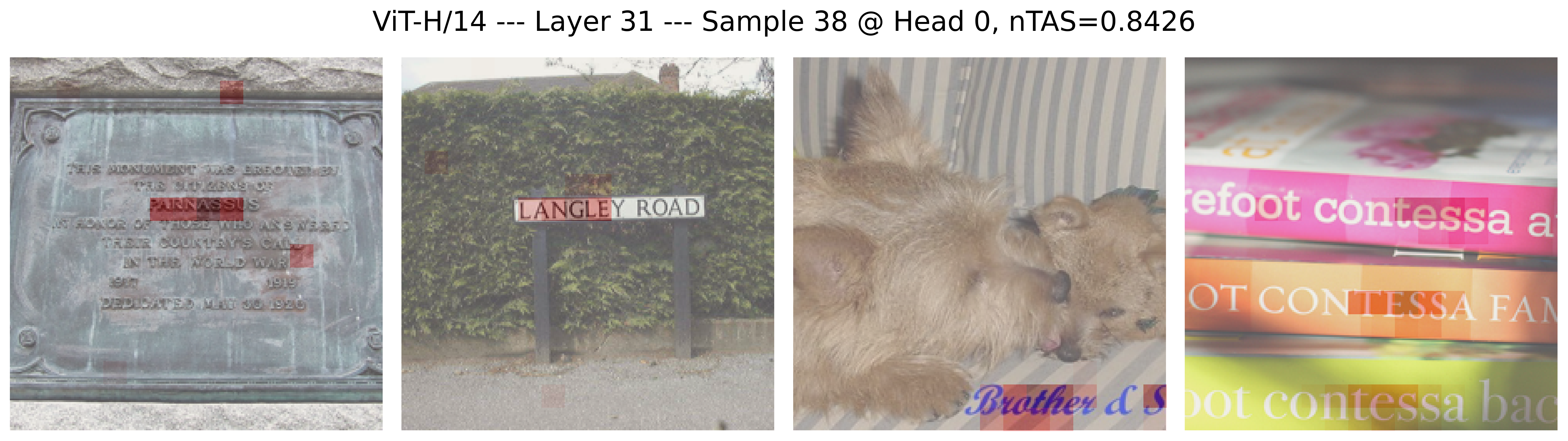} & 
        \includegraphics[width=0.495\linewidth]{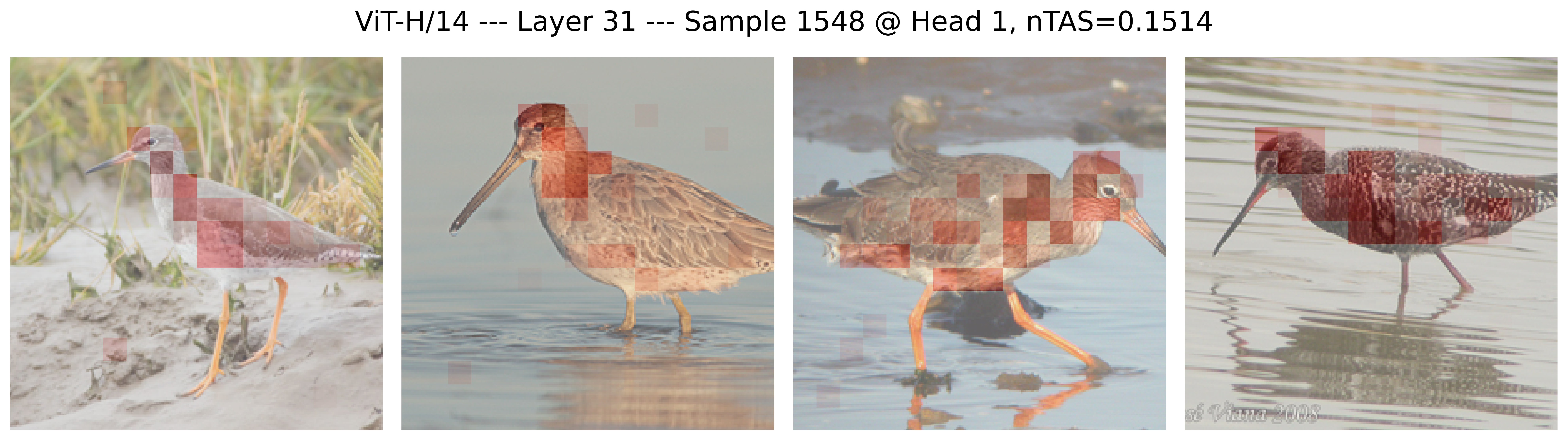} \\
        \includegraphics[width=0.495\linewidth]{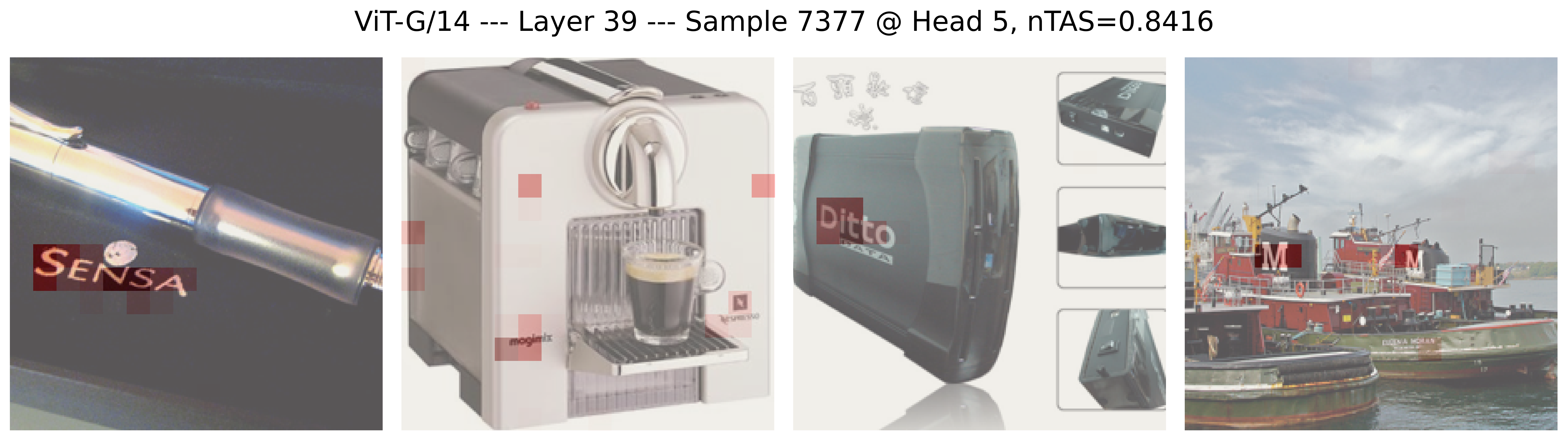} & 
        \includegraphics[width=0.495\linewidth]{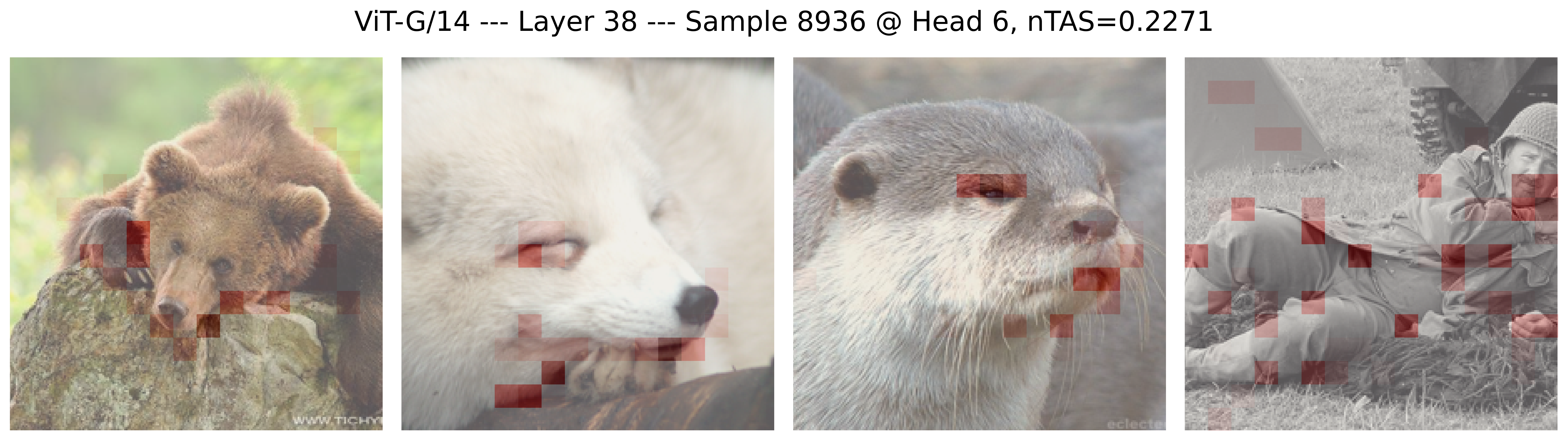} \\
        \includegraphics[width=0.495\linewidth]{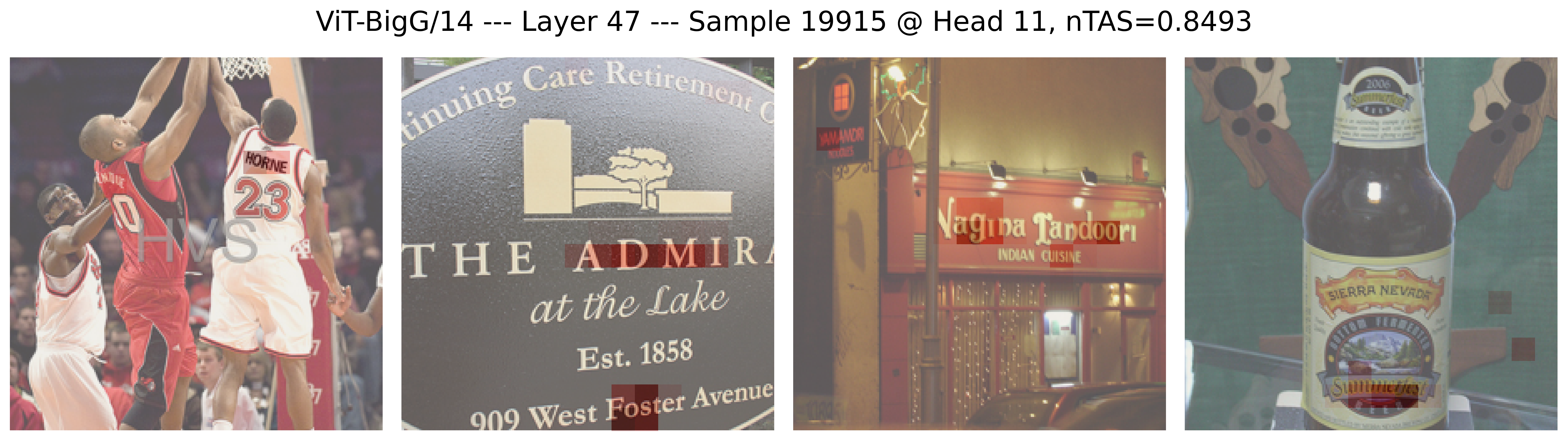} & 
        \includegraphics[width=0.495\linewidth]{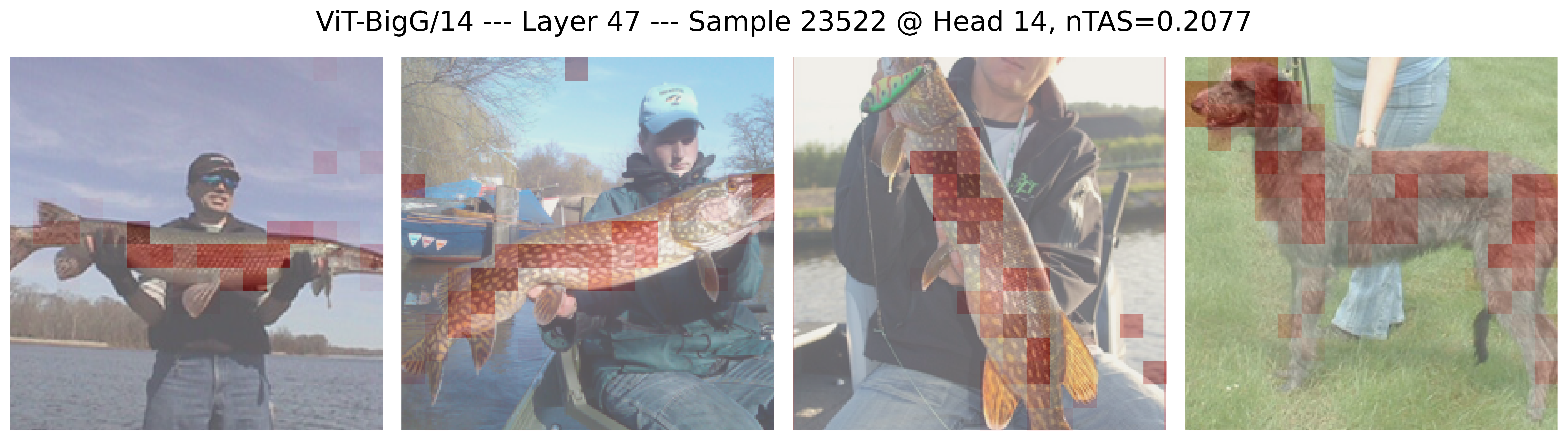} \\
    \end{tabular}
    }
    \end{center}
    \caption{\textbf{Concept Localization.} Attribution map of selected random concept vectors with high text focus indicated by a high nTAS (Left), and with low text focus indicated by a low nTAS (Right). High attribution-score patches are indicated by a red gradient. }
    \label{fig:viz}
\end{figure}

\cref{fig:alignment} shows that nTAS is robust against unaligned concept directions, as attributing an off-concept vector produces a diffuse, noisy attribution map without semantic focus. Beyond this null-case check, we further validate the extracted circuits and the capacity of the sampling and ranking process in~\cref{fig:viz}, which shows the attribution maps generated by selected concept vectors on their activating images: when contrasting high-scoring nTAS cohorts with their low-scoring counterparts, a positive association is observed between nTAS magnitude and the degree of lexical focus. This supports our design of nTAS for TA-vulnerable circuit mining.
\subsection{Ablation and Variation Studies}
We evaluate the stability of our method by repeating the object classification experiments with 4 random seeds and expansion ratios of $4,8,16,32,64$.
\begin{table}[tb]

\centering
\caption{\textbf{Robustness to random seed variations.} We report the mean and standard deviation for object classification accuracy (OCA) \% and text confusion rate (TCR) \% at an expansion ratio of 16. The low standard deviations demonstrate the stability of our method.}
\label{tab:seed_robustness}
\resizebox{0.8\textwidth}{!}{%
\setlength{\tabcolsep}{8pt}
\begin{tabular}{@{}llcccc@{}}
\toprule
\textbf{Model} & \textbf{Metric} & \textbf{RTA-100} & \textbf{Disentangling} & \textbf{PAINT} & \textbf{IN-100-Text} \\
\midrule
\multirow{2}{*}{ViT-B/16} 
 & OCA & $68.7 \pm 0.0$ & $88.3 \pm 0.0$ & $73.8 \pm 0.0$ & $74.2 \pm 0.0$ \\
 & TCR  & $12.6 \pm 0.0$ & $11.7 \pm 0.0$ & $16.5 \pm 0.0$ & $7.1 \pm 0.0$ \\
\midrule
\multirow{2}{*}{ViT-L/14} 
 & OCA & $68.8 \pm 0.1$ & $68.3 \pm 0.0$ & $68.9 \pm 0.0$ & $74.7 \pm 0.3$ \\
 & TCR  & $21.3 \pm 0.2$ & $31.1 \pm 0.0$ & $22.3 \pm 0.0$ & $12.2 \pm 0.3$ \\
\midrule
\multirow{2}{*}{ViT-H/14} 
 & OCA & $76.2 \pm 0.1$ & $83.2 \pm 1.9$ & $77.2 \pm 2.9$ & $79.2 \pm 0.1$ \\
 & TCR  & $14.1 \pm 0.6$ & $16.3 \pm 1.9$ & $13.6 \pm 1.9$ & $9.2 \pm 0.5$ \\
\midrule
\multirow{2}{*}{ViT-g/14} 
 & OCA & $68.8 \pm 0.0$ & $83.2 \pm 3.1$ & $75.7 \pm 0.0$ & $76.4 \pm 0.1$ \\
 & TCR  & $23.4 \pm 0.1$ & $16.3 \pm 3.1$ & $18.2 \pm 0.5$ & $12.6 \pm 0.3$ \\
\midrule
\multirow{2}{*}{ViT-bigG/14} 
 & OCA & $75.4 \pm 0.2$ & $70.6 \pm 2.6$ & $75.0 \pm 3.7$ & $80.0 \pm 0.4$ \\
 & TCR  & $15.7 \pm 0.6$ & $28.9 \pm 2.6$ & $12.9 \pm 2.0$ & $9.6 \pm 0.6$ \\
\bottomrule
\end{tabular}%
}
\end{table}
\subsubsection{Consistency over Random States.}
To showcase the stability of our method, we repeat the object classification experiment with 4 random seeds and report the results in \cref{tab:seed_robustness}. Our method consistently extracts faithful lexical circuits and effectively promotes TA robustness across ViT variants and datasets, as evidenced by the low standard deviation of OCA and TCR.
\subsection{Influence of the Number of Samples.}
\begin{figure}[tbp]
    \centering
    \includegraphics[width=0.8\linewidth]{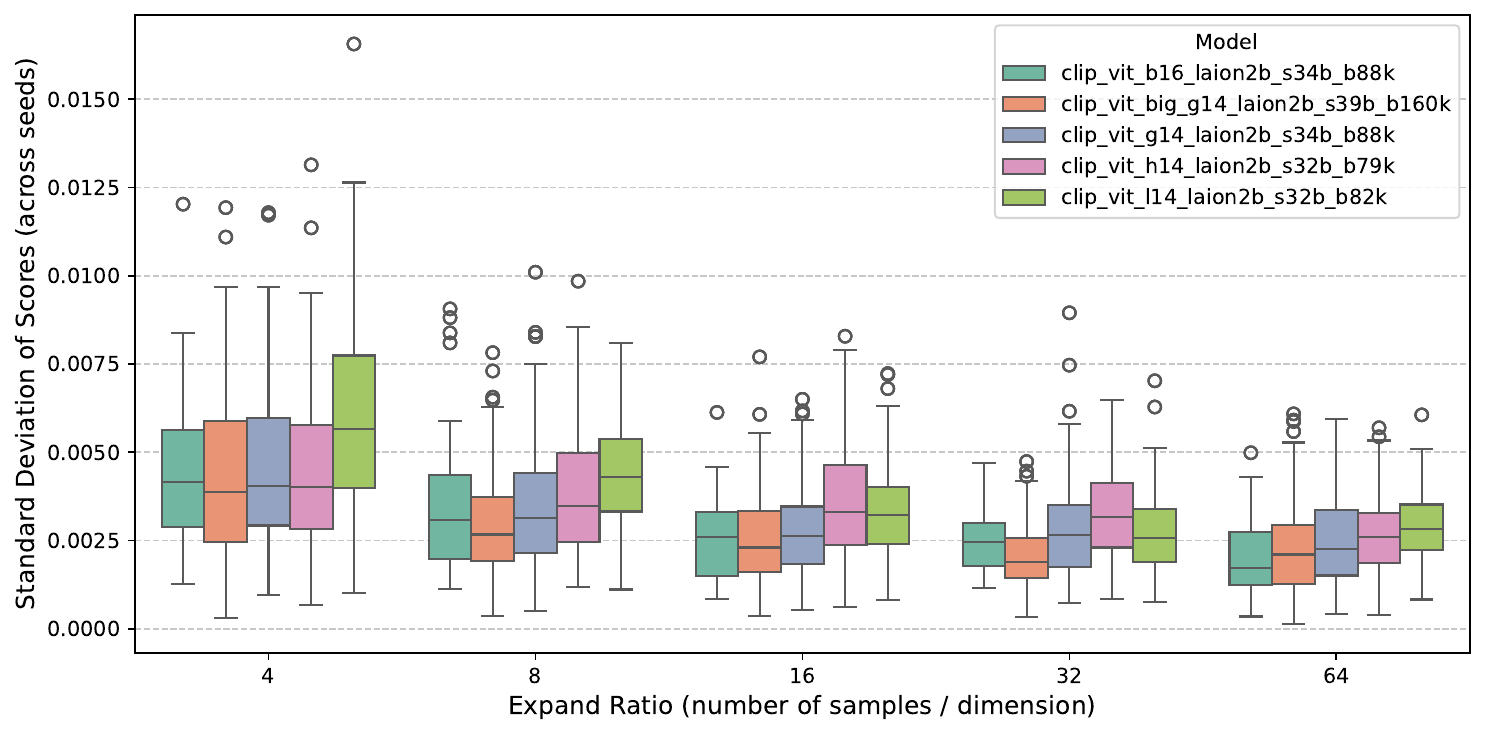}
    \caption{Impact of the number of samples (parametrized by expansion ratio) on the method stability (in the standard deviation of the module score across random seeds). Our method stabilizes at 8 samples per dimension across ViT sizes and continues to stabilize as the number of samples increases.}
    \label{fig:sample_vs_std}
\end{figure}

As shown in \cref{fig:sample_vs_std}, the standard deviation of the module score across random states decreases as the number of samples increases, which demonstrates that our training-free model explanation stabilizes with a large number of samples. 
\section{Conclusion}
We tackle the challenge of interpreting hidden concept directions in ViT intermediate states using unlabeled images and efficiently extract causal circuits responsible for typographic-attack vulnerability. Our method treats concept discovery as a stochastic lottery within the lower-dimensional MHSA head subspace, grounded in the Linear Representation Hypothesis~\cite{park2024linear, elhage2022superposition} and the Lottery Ticket Hypothesis~\cite{frankle2018the}, and uses a gradient-based attribution score to mine the lexical circuits. The proposed circuit mining pipeline requires only a diverse, unlabeled image dataset and minimal validation data for threshold hyperparameter search.  It improves robustness to typographic attacks across model scales and datasets, surpassing prior supervised and training-free defenses. Extension of our method to LVLM demonstrates the ability to improve the VQA accuracy of LVLMs under lexical distraction. Finally, the qualitative analysis of the neuron's activations verifies the causal relationship between the attention heads' focus on text and their correspondence in the lexical circuits. We hope this research will advance the frontier of mechanistic solutions to VLM robustness.
\section*{Acknowledgment}
This work is supported in part by the US National Science Foundation (NSF) under grants IIS-2538206, IIS-2529378, IIS-2500341, CCF-2217071, CNS-2213700, and OAC-2530655. Any recommendations expressed in this material are those of the authors and do not necessarily reflect the views of NSF.

%% file: sec/a_appendix.tex
\appendix
\section{Appendix}
\subsection{Dataset Details}
\begin{figure}
    \centering
    \includegraphics[width=1\linewidth]{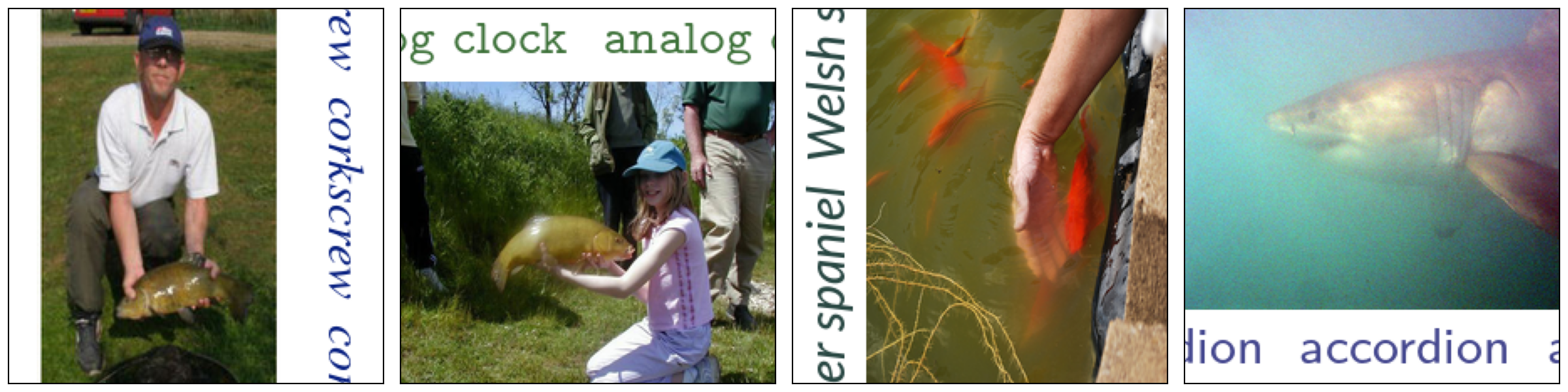}
    \caption{\textbf{Example of text-augmented images.} Texts with various colors and fonts occur at each border with equal likelihood.}
    \label{fig:augmentation}
\end{figure}

\noindent \textbf{Circuit Mining Dataset.} Our training-free concept mining process takes a text-injected image dataset without labels. In our experiments, we use a uniformly sampled $0.1\%$ subset of ImageNet~\cite{imnet} as the base dataset. The texts are randomly rendered in various colors and fonts at the image margins, spanning $20\%$ of the image width, as shown in \cref{fig:augmentation}. The proposed method requires only the text-injected image and the text location, provided by the augmentation.

\begin{figure}[t]
    \centering
    \includegraphics[width=1\linewidth]{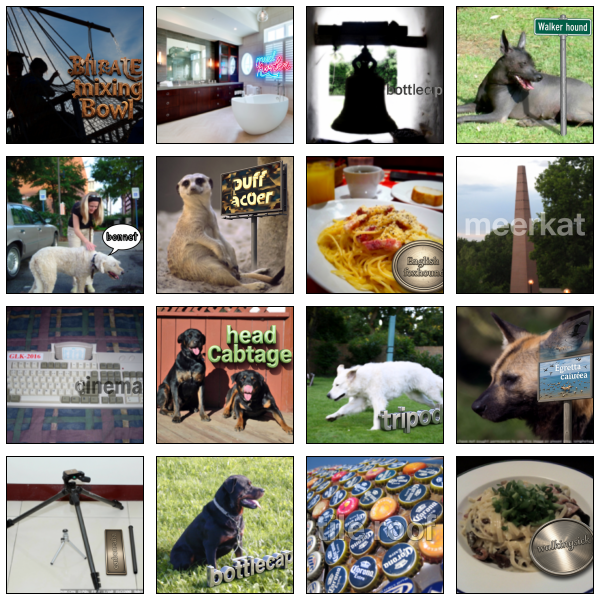}
    \caption{\textbf{Example data from IN-100-Text.} }
    \label{fig:data_sample}
\end{figure}

\noindent \textbf{Evaluation Datasets.} We evaluate the defense methods on the following datasets:
\begin{itemize}
    \item \textbf{RTA-100}~\cite{denseprefix} consists of 100 categories and a 1000-item mixture of synthetic and real-world typographic attack images. The images are drawn from 10 regular image datasets, 2 typographic attack datasets, and additional images collected from the literature. 
    \item \textbf{Disentangling}~\cite{disentangle} consists of 19 categories and 171 images, constructed with daily objects and attack text written on sticky notes. 
    \item \textbf{PAINT}~\cite{ilharco2022patching} contains 110 images of daily objects and attack-text sticky notes.
    \item \textbf{IN-100-Text} is constructed from CMC's dataset~\cite{clane_im100}, a 100-class subset of ImageNet~\cite{imnet}. We use the validation set of 5000 images to augment the data with Qwen-Image-Edit~\cite{wu2025qwen}. We consider seven common text types that often co-occur with objects across varying scenarios and are visualized in ways not covered by previous datasets: neon sign, comic chat balloon, street sign, billboard, engraved metal plate, sculpted 3D CGI text, and obvious watermark. The prompt for generating the text-injected images is shown in \cref{fig:qwen-prompt}, and a visualized subset of all images in the generated dataset is given in \cref{fig:data_sample}
    \item \textbf{ImageNet-100}~\cite{shekhar2021imagenet100} is a subset of ImageNet consisting of 100 classes. We use it to evaluate defense trade-offs in terms of standard image classification accuracy relative to other methods. 
\end{itemize}
\begin{figure}[t]
    \centering
    \begin{promptbox}[title={User Prompt}]
    <image> Add a \{text\_type\} displaying the text "\{text\_word\}" next to the \{gt\_text\}.
    \end{promptbox}
    \caption{\textbf{Prompt for Typographic Image Editing.} text\_type refers to one of the seven text types we defined. text\_word is chosen randomly from the class label set, excluding the ground truth label of the input image. gt\_text refers to the ground truth label of the input image.}
    \label{fig:qwen-prompt}
\end{figure}

\subsection{Derivation of $P_{\text{success}}$}
\subsubsection{Bounding Interference via Concentration of Measure.}
We establish a high-probability bound on the maximum polysemantic interference
across $N$ sequence patches. Let the random probe $\mathbf{u}$ be drawn from an
isotropic Gaussian scaled by the head dimension:
\begin{equation}
\mathbf{u} \sim \mathcal{N}\left(\mathbf{0}, \frac{1}{d_{\text{head}}}\mathbf{I}\right).
\end{equation}

For patch $i$, let $\boldsymbol{\xi}_i \in \mathbb{R}^{d_{\text{head}}}$ be the
deterministic interference vector orthogonal to $\mathbf{c}_{\text{target}}$. Its
projection onto the probe is a zero-mean Gaussian:
\begin{equation}
I_{\mathbf{u},i} = \langle \boldsymbol{\xi}_i, \mathbf{u} \rangle
\sim \mathcal{N}\!\left(0, \frac{\|\boldsymbol{\xi}_i\|^2}{d_{\text{head}}}\right).
\end{equation}

For a Gaussian $Z \sim \mathcal{N}(0,\sigma^2)$, the sub-Gaussian tail bound gives
$\mathbb{P}(|Z| \ge t) \le 2\exp(-t^2/2\sigma^2)$. Applied to the interference
projection and combined with the union bound over all $N$ patches, with
$\|\boldsymbol{\xi}_{\text{max}}\| = \max_i \|\boldsymbol{\xi}_i\|$:
\begin{equation}
\mathbb{P}\!\left(\max_{i \in \{1\dots N\}} |I_{\mathbf{u},i}| \ge t\right)
\le 2N \exp\!\left(-\frac{t^2 d_{\text{head}}}{2\|\boldsymbol{\xi}_{\text{max}}\|^2}\right).
\end{equation}

We set the threshold to
$t = \|\boldsymbol{\xi}_{\text{max}}\| \sqrt{\frac{2(1+\delta)\log N}{d_{\text{head}}}}$
for a slack parameter $\delta > 0$. The exponent becomes $-(1+\delta)\log N$, so
the exponential contributes $N^{-(1+\delta)}$ and the union-bound factor $N$ is
overpowered:
\begin{equation}
\mathbb{P}\!\left(\max_{i} |I_{\mathbf{u},i}|
\ge \|\boldsymbol{\xi}_{\text{max}}\| \sqrt{\frac{2(1+\delta)\log N}{d_{\text{head}}}}\right)
\le 2N \cdot N^{-(1+\delta)} = 2N^{-\delta}.
\end{equation}

The failure probability $2N^{-\delta}$ decays polynomially in $N$ for any
$\delta > 0$, with $\delta$ trading bound tightness ($2N^{-\delta}$) against the
separation threshold ($\propto \sqrt{1+\delta}$). We set $\delta = 1$ to balance
the two, yielding a failure probability of $2/N$ and the high-probability bound
\begin{equation}
\label{eq:condition-app}
\max_{i \in \{1\dots N\}} |I_{\mathbf{u},i}|
\le \|\boldsymbol{\xi}_{\text{max}}\| \sqrt{\frac{4\log N}{d_{\text{head}}}}.
\end{equation}
\subsubsection{Derivation of the Single Success Probability ($p$).}
Next, we derive the probability $p$ that a single random probe $\mathbf{u}$
resolves the weakest on-concept patch $k = \arg\min_{i \in \text{on}} \alpha_i$ above the background interference. Since every on-concept patch carries the concept at least as strongly as $\alpha_k$, resolving patch $k$ resolves all on-concept patches simultaneously.
The alignment of the probe with the concept direction is
$A_{\mathbf{u}} = \langle \mathbf{c}_{\text{target}}, \mathbf{u} \rangle$, with
\begin{equation}
A_{\mathbf{u}} \sim \mathcal{N}\!\left(0, \frac{1}{d_{\text{head}}}\right).
\end{equation}
The signal of the weakest on-concept patch is
$S_{\mathbf{u},k} = \alpha_k A_{\mathbf{u}}$, distributed as
\begin{equation}
S_{\mathbf{u},k} \sim \mathcal{N}\!\left(0, \frac{\alpha_k^2}{d_{\text{head}}}\right).
\end{equation}
To utilize standard normal bounds, we scale $S_{\mathbf{u},k}$ to a standard normal variable $Z \sim \mathcal{N}(0, 1)$:
\begin{equation}
Z = \frac{S_{\mathbf{u},k} \sqrt{d_{\text{head}}}}{\alpha_k} \sim \mathcal{N}(0,1).
\end{equation}

For successful separation, the signal of patch $k$ must exceed the sum of $\tau$ and the maximum interference bound derived in Section 3.4:
\begin{equation}
S_{\mathbf{u},k} > \|\boldsymbol{\xi}_{\text{max}}\| \sqrt{\frac{4 \log N}{d_{\text{head}}}} + \tau .
\end{equation}

Multiplying both sides by $\sqrt{d_{\text{head}}}$ translates this condition into our standard normal variable $Z$:
\begin{equation}
Z > \frac{\|\boldsymbol{\xi}_{\text{max}}\| \sqrt{4 \log N} + \tau \sqrt{d_{\text{head}}}}{\alpha_k} .
\end{equation}

Using the Gaussian tail approximation $\mathbb{P}(Z > z) \approx \exp(-z^2/2)$ for large $z$, we obtain the estimated probability $p$ of drawing a successful separation vector:
\begin{equation}
p \approx \exp\left( - \frac{\left( \|\boldsymbol{\xi}_{\text{max}}\| \sqrt{4 \log N} + \tau \sqrt{d_{\text{head}}} \right)^2}{2\alpha_k^2} \right)
\end{equation}

\subsubsection{Derivation of the Required Sample Size ($K$).}
Finally, we calculate the number of random vectors $K$ required in our Stochastic Lottery to guarantee finding at least one clean feature map with a desired confidence level, $P_{\text{success}}$.

Given that each of the $K$ probes is drawn independently, the probability that a single probe fails to isolate the concept is $(1 - p)$. The probability that \textbf{all} $K$ probes fail is the product of their individual failure probabilities:
\begin{equation}
\mathbb{P}(\text{all fail}) = (1 - p)^K
\end{equation}

The probability of obtaining at least one successful probe is the complement of total failure. We require this to be greater than or equal to $P_{\text{success}}$:
\begin{equation}
1 - (1 - p)^K \ge P_{\text{success}}
\end{equation}

We rearrange this inequality to solve for $K$:
\begin{equation}
(1 - p)^K \le 1 - P_{\text{success}}
\end{equation}

Taking the natural logarithm of both sides yields:
\begin{equation}
K \log(1 - p) \le \log(1 - P_{\text{success}})
\end{equation}

Because $p$ is a probability between 0 and 1, the term $(1 - p)$ is less than 1, making its logarithm strictly negative. Dividing both sides by a negative number flips the inequality sign, resulting in the final lower bound for $K$:
\begin{equation}
K \ge \frac{\log(1 - P_{\text{success}})}{\log(1 - p)}
\end{equation}

This completes the derivation, justifying the benefit of dimension reduction from operating within the $d_{\text{head}}$ subspace.

\subsection{nTAS Distribution over ViT Model Depth}
\begin{figure}[t]
    \centering
    \includegraphics[width=0.8\linewidth]{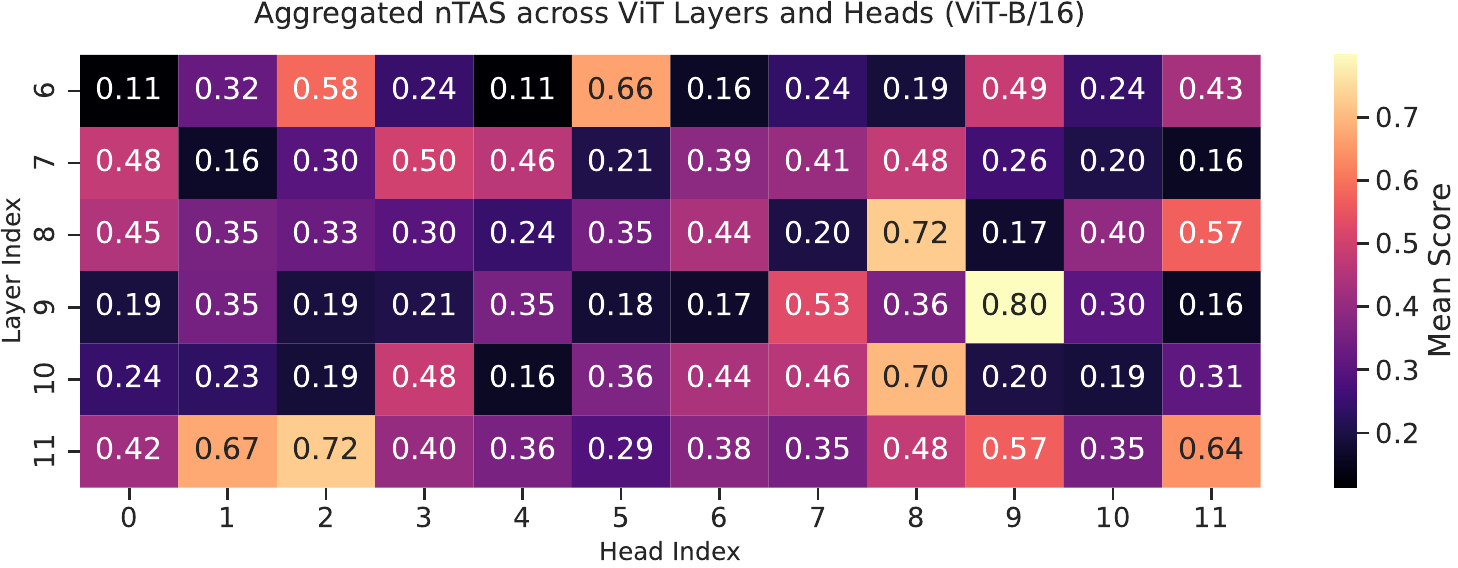}
    \includegraphics[width=1\linewidth]{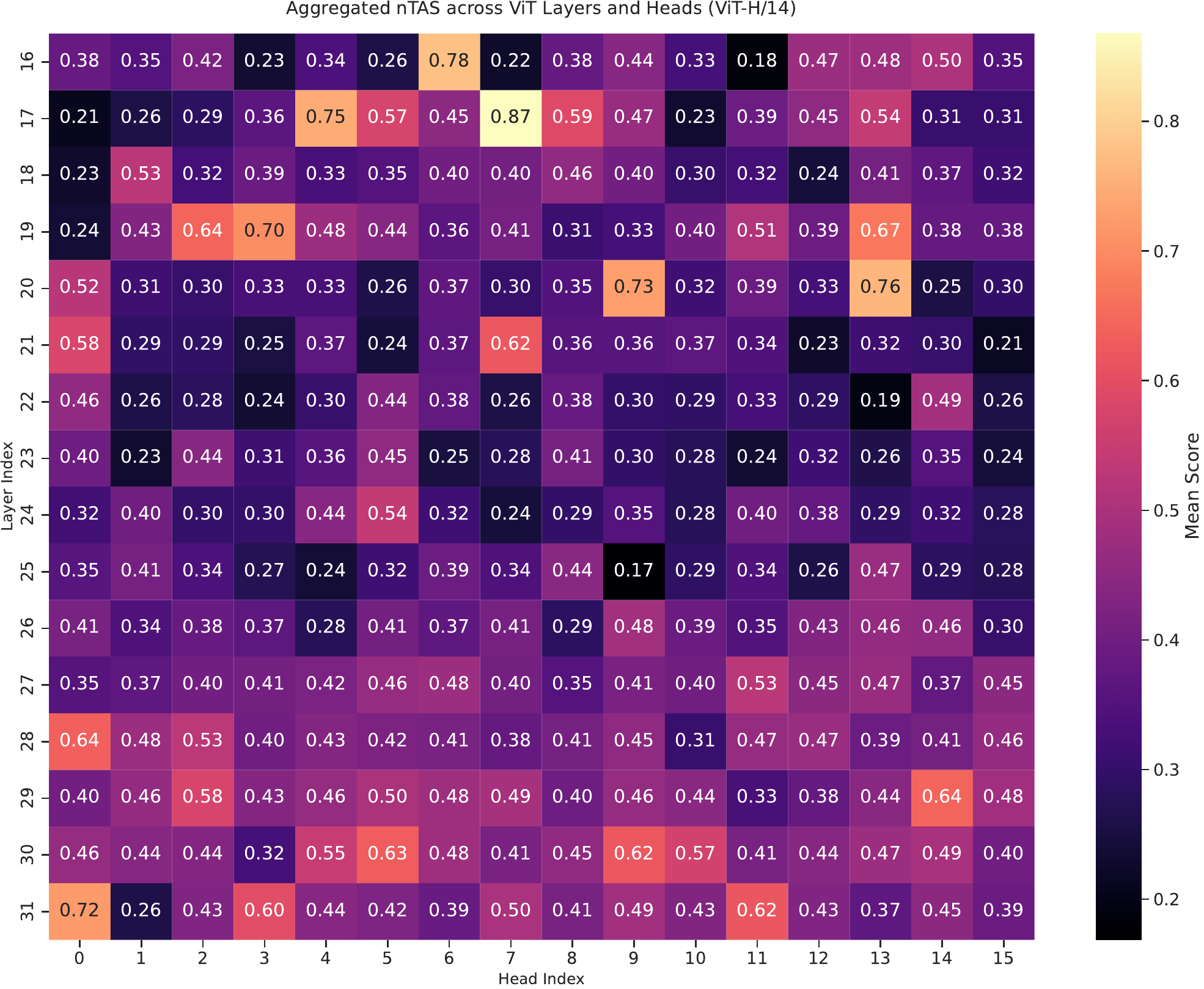}
    \caption{\textbf{Mean nTAS of Attention Heads across ViT layers.}}
    \label{fig:ntas_by_head}
\end{figure}

In \cref{fig:ntas_by_head}, we show the per-attention-module nTAS across the last half of the ViT-B/16 and the ViT-H/14 models. Our method can produce distinct lexical-focused attention modules across ViT layers, as highlighted in \cref{fig:ntas_by_head}. The comparison between ViT-B/16 and ViT-H/14 reveals a potential discrepancy in model-level internal behavior related to model size: ViT-B/16 has more evenly distributed lexical attention modules, whereas ViT-H/14 exhibits a "dark zone" from layer 22 to layer 26. Such an observation sparks further mechanistic interpretability research into shifts in ViT model behavior with respect to model size. 

\subsection{More Results on General Performance Trade-off}
As shown in \cref{tab:tradeoff}, our method consistently yields competitive general classification results compared with the methods we compare against. The supervised optimization-based Defense-Prefix tuning yields the least trade-off in overall performance, as the optimization process preserves model behavior on normal inputs. Notably, our method achieves comparable or better-than-supervised general classification accuracy across two of the five ViT variants tested. 

\begin{table}[tbp]
\centering
\setlength{\tabcolsep}{5pt}
\caption{\textbf{Performance Tradeoff.} We present the performance trade-off of the compared methods in terms of classification accuracy on the ImageNet-100 validation set. }
\label{tab:tradeoff}
\resizebox{\linewidth}{!}{
\begin{tabular}{lcccccc}
\toprule
Method & ViT-B/16 & ViT-L/14 & ViT-H/14 & ViT-g/14 & ViT-bigG/14 & Average \\
\midrule
Vanilla ViT  & 74.8        & 80.0          & 83.8        & 83.8        & 85.3        & 81.5 \\
Defense-Prefix & 75.4 (\textbf{+0.6}) & 79.8 (\textbf{-0.2}) & 83.4 (\textbf{-0.4}) & 83.2 (-0.6) & 85.4 (\textbf{+0.1}) & 81.4 (\textbf{-0.14}) \\
Dyslexify    & 75.3 (+0.5) & 79.5 (-0.5) & 83.4 (\textbf{-0.4}) & 83.0 (-0.8)   & 85.0 (-0.3)   & 81.3 (-0.24) \\
Dyslexify* & 75.0 (+0.2)   & 79.5 (-0.5) & 83.4 (\textbf{-0.4}) & 82.6 (-1.2) & 84.7 (-0.6) & 81.0 (-0.54) \\
Ours (nTAS)  & 74.2 (-0.6) & 79.8 (\textbf{-0.2}) & 82.9 (-0.9) & 83.3 (\textbf{-0.5}) & 84.6 (-0.7) & 81.0 (-0.54) \\
\bottomrule
\end{tabular}
}
\end{table}

\begin{figure}[t]
    \centering
    \includegraphics[width=1\linewidth]{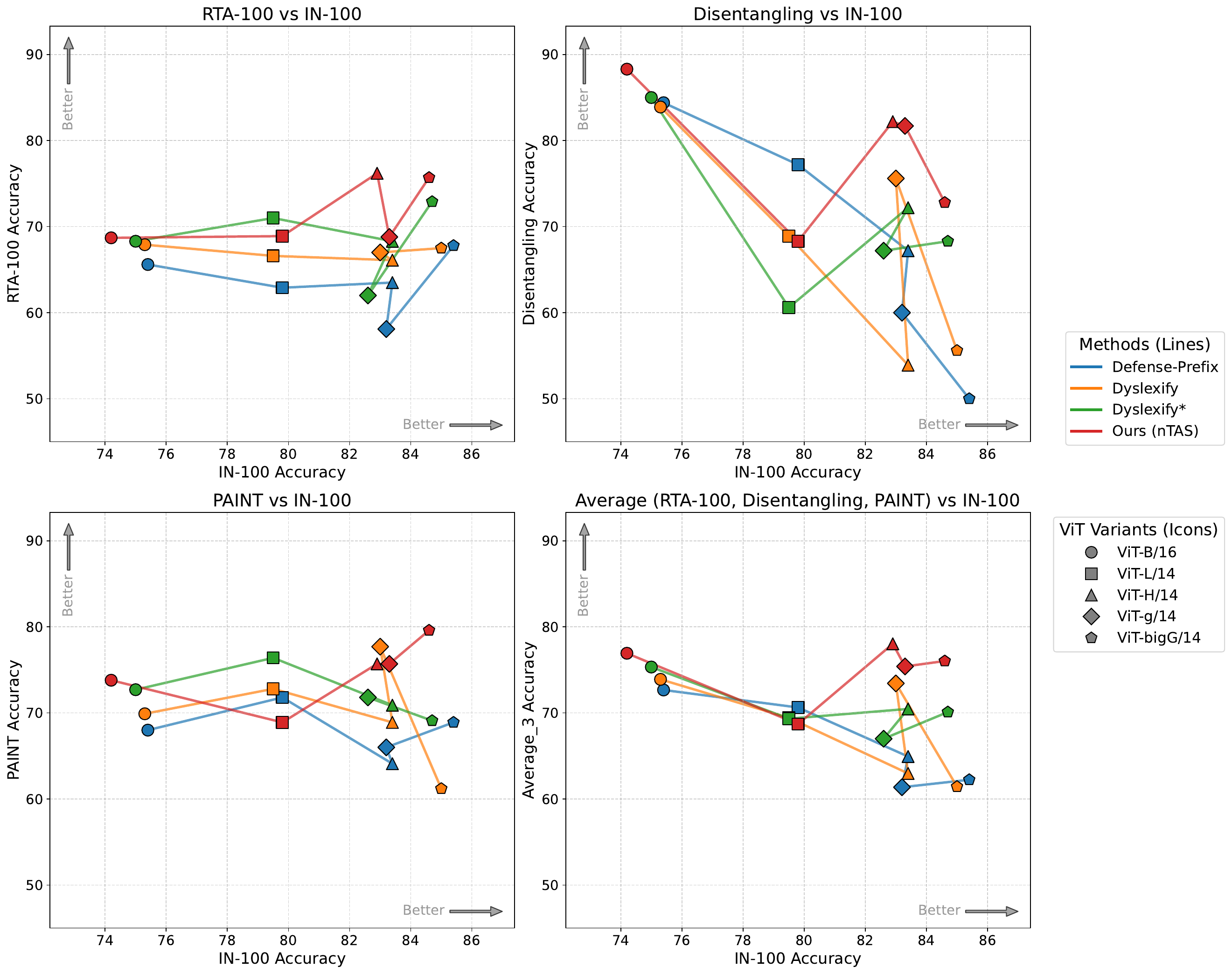}
    \caption{\textbf{Classification Accuracy on Typographic-Attack Datasets and ImageNet-100 Dataset.} The joint plot shows the overall robustness trade-off of each tested method. Our method consistently produces strong robustness with minimal trade-off, especially in large and complex ViT variants.}
    \label{fig:tradoff}
\end{figure}

\cref{fig:tradoff} shows the robustness-capacity tradeoff by plotting the typographic-attack dataset accuracies and the ImageNet-100 dataset accuracies. Our method consistently yields the best results among large ViT models, benefiting from the model-agnostic nature of the top-down attribution-based circuit mining process. 
